\documentclass[lettersize,journal]{IEEEtran}
\usepackage{amsmath,amsfonts}
\usepackage{amssymb}
\usepackage{algorithmic}
\usepackage{algorithm}
\usepackage{array}
\usepackage{booktabs}
\usepackage{dsfont}
\usepackage[caption=false,font=normalsize,labelfont=sf,textfont=sf]{subfig}
\usepackage{textcomp}
\usepackage{stfloats}
\usepackage{url}
\usepackage{verbatim}
\usepackage{graphicx}
\usepackage{cite}
\usepackage{xspace}
\newcommand{\ie}{{\emph{i.e.}}\xspace}
\newcommand{\eg}{{\emph{e.g.}}\xspace}

\newcommand{\etal}{{\emph{et al.}}\xspace}
\usepackage[pagebackref=false,breaklinks=true,letterpaper=true,colorlinks,linkcolor=red,bookmarks=false]{hyperref}
\newcommand{\thickhline}{%
    \noalign {\ifnum 0=`}\fi \hrule height 1pt
    \futurelet \reserved@a \@xhline
}
\usepackage{multirow}
\usepackage{multicol}
\usepackage{color}

\usepackage{bbding}
\usepackage{pifont}
\usepackage{booktabs}
\usepackage{placeins}
\usepackage{tabularx}
\usepackage{caption}
\usepackage{color,xcolor}
\usepackage{stfloats}
\usepackage{wasysym}
\usepackage{xspace}
\usepackage{bbm}
\newcommand{\cmark}{\ding{51}\xspace}%
\newcommand{\xmarkg}{\textcolor{lightgray}{\ding{55}}\xspace}%
\definecolor{myblue}{RGB}{0,120,191}
\definecolor{myred}{RGB}{204,64,84}
\definecolor{mygreen}{RGB}{19,138,7}

\newcommand{{\ourmodel}}{VGSG}

\begin{document}

\title{VGSG: Vision-Guided Semantic-Group Network for Text-based Person Search}
\author{Shuting He,
        Hao Luo,
        Wei Jiang,
        Xudong, Jiang~\IEEEmembership{IEEE Fellow},
        Henghui Ding
\thanks{Shuting He, Xudong Jiang, and Henghui Ding are with Nanyang Technological University (NTU), Singapore 639798
(e-mail: shuting.he@ntu.edu.sg; exdjiang@ntu.edu.sg; henghui.ding@gmail.com).}
\thanks{Hao Luo and Wei Jiang are with Zhejiang University,
Hangzhou 310027, China (e-mail:  haoluocsc@zju.edu.cn; jiangwei\_zju@zju.edu.cn).}

}

\markboth{IEEE TRANSACTIONS ON IMAGE PROCESSING}%
{Shell \MakeLowercase{\textit{et al.}}: A Sample Article Using IEEEtran.cls for IEEE Journals}

\maketitle

\begin{abstract}
Text-based Person Search (TBPS) aims to retrieve images of target pedestrian indicated by textual descriptions. It is essential for TBPS to extract fine-grained local features and align them crossing modality. Existing methods utilize external tools or heavy cross-modal interaction to achieve explicit alignment of cross-modal fine-grained features, which is inefficient and time-consuming. In this work, we propose a Vision-Guided Semantic-Group Network (VGSG) for text-based person search to extract well-aligned fine-grained visual and textual features. In the proposed VGSG, we develop a Semantic-Group Textual Learning (SGTL) module and a Vision-guided Knowledge Transfer (VGKT) module to extract textual local features under the guidance of visual local clues. In SGTL, in order to obtain the local textual representation, we group textual features from the channel dimension based on the semantic cues of language expression, which encourages similar semantic patterns to be grouped implicitly without external tools. In VGKT, a vision-guided attention is employed to extract visual-related textual features, which are inherently aligned with visual cues and termed vision-guided textual features. Furthermore, we design a relational knowledge transfer, including a vision-language similarity transfer and a class probability transfer, to adaptively propagate information of the vision-guided textual features to semantic-group textual features. With the help of relational knowledge transfer, VGKT is capable of aligning semantic-group textual features with corresponding visual features without external tools and complex pairwise interaction. Experimental results on two challenging benchmarks demonstrate its superiority over state-of-the-art methods.
\end{abstract}


\section{Introduction}\label{sec:introduction}

\IEEEPARstart{T}{ext}-based person search (TBPS)~\cite{li2017person} is a challenging cross-modal retrieval task. It aims to retrieve images of the target person indicated by a given textual description. Compared to the query image for image-based person search, textual description is much easier accessible and more flexible in real-world applications. Recently, text-based person search has gained more and more attention~\cite{lgur,lapscore,textreid,ssan,wang2021mgel,zheng2020hierarchical,gao2021contextual,chen2021cmka,zheng2020dual}.
After comprehensively revisiting recent state-of-the-art methods, we find that there are two critical issues in the field of text-based person search. Firstly, learning fine-grained local features is critical for TBPS because samples from different classes may have only small inter-class variation~\cite{ssan,jing2020pose,wang2021mgel, farooq2021axm,zheng2020hierarchical,chen2021tipcb}. Secondly, alleviating misalignment of fine-grained cross-modal features is important since there is a large modality gap between vision and language~\cite{lgur,li2017identity,VLT,chen2021cmka,aggarwal2020cmaam}.
It is important for text-based person search to learn fine-grained local features and align them over crossing modality for retrieval.
\begin{figure*}[t]
	\centering
	\includegraphics[width = \textwidth]{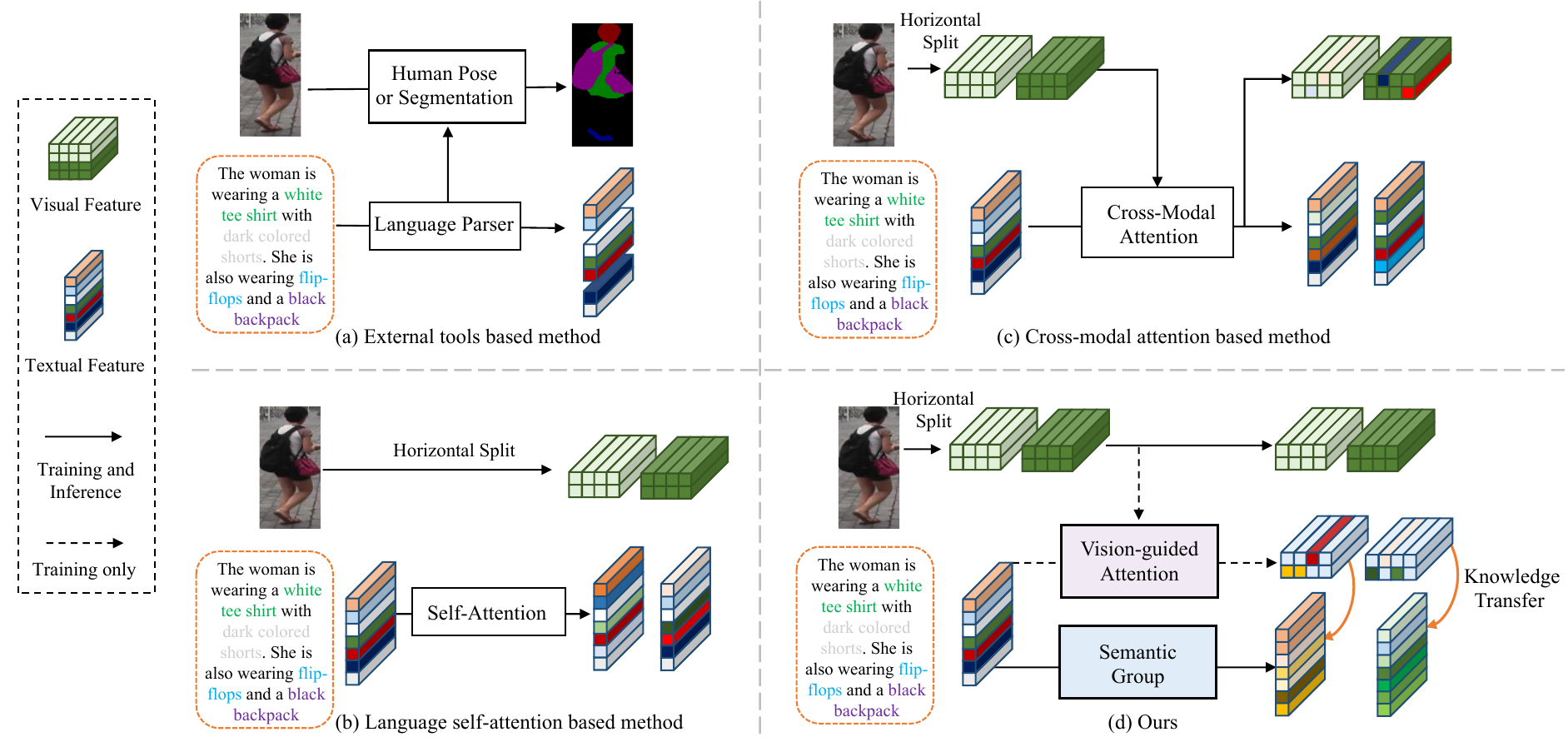}
	\caption{\small{Four methods to extract aligned fine-grained local visual and textual features. (a) External tool based method. (b) Language self-attention based method. (c) Cross-modal attention based method. (d) Ours proposed CLIP-driven Text-based Person Search (\ourmodel).}}
	\label{fig:motivation}
\end{figure*}

Previous methods that aim at alleviating these challenges can be categorized into three mainstreams, \ie, external tool based methods, language self-attention based methods, and cross-modal attention based methods, as shown in \figurename~\ref{fig:motivation} (a) to (c).
Firstly, external tool based methods directly apply external cues, \eg, language parser NLTK~\cite{loper2002nltk} or segmentation~\cite{wang2020vitaa,jing2020pose,aggarwal2020cmaam} to achieve explicit alignment of the cross-modal fine-grained features. 
For example, VITAA~\cite{wang2020vitaa} utilizes NLTK~\cite{loper2002nltk} to detect language noun phrases as attributes, then decouples the visual feature space into sub-spaces according to an auxiliary attribute segmentation layer (see \figurename~\ref{fig:motivation}(a)). However, such an approach largely relies on the performance of the external models and is susceptible to noise. In addition, they are time-consuming and inefficient due to using extra models.
Secondly, language self-attention based methods~\cite{ssan,farooq2021axm} exploit language self-attention mechanisms to extract the informative and emphasized word(s) as fine-grained local textual cues. For local visual features, they perform uniformly partition to extract local textual features inspired by image-based ReID~\cite{sun2018beyond, MGN}.
For example, as shown in \figurename~\ref{fig:motivation}(b), SSAN~\cite{ssan} designs a self-aligned part module for local textual features extraction, but their linguistic partition considers only the language information itself while neglecting the image information. Hence, they cannot well align the extracted local textual features with a specific visual part. 
Thirdly, as shown \figurename~\ref{fig:motivation}(c), cross-modal attention based methods utilize cross-modal attention~\cite{li2017person,li2017identity,gao2021contextual} to construct interaction between body parts in images and words in texts. However, this paradigm needs repeatedly align the features for each image-text pair via the cross-modal attention module, leading to high computational cost during the inference stage. Especially, the computational complexity is overwhelming when handling large-scale datasets.

In light of the above observations, we propose a Vision-Guided Semantic-Group (\ourmodel) network for Text-based Person Search to explore implicit alignment of fine-grained features efficiently, as shown in \figurename~\ref{fig:motivation}(d).
First, inspired by the superior performance of CLIP~\cite{clip} in the field of vision-language tasks, we construct a strong CLIP-driven baseline framework to mitigate the misalignment of fine-grained cross-modal features owing to the powerful vision-language alignment capability of CLIP.
Then, we propose a Semantic-Group Textual Learning (SGTL) module to group textual features from channel dimension relying on the semantic distribution of language expression to excavate local textual cues.
We hypothesize that channels of textual features can be grouped according to the semantic distribution corresponding to a certain type of visual pattern~\cite{zhang2016picking}. Specifically, 
we feed channel-group textual features into a transformer block to explore the semantic correlation between words in a global context. In this way, textual features can implicitly encourage similar semantic patterns to be grouped together without utilizing extra tools and we term the final output the semantic-group textual features.

However, it is challenging to ensure semantic-group textual features are aligned with corresponding visual concepts without the assistance of semantic annotations.
A large number of image-based ReID methods~\cite{sun2018beyond,MGN} show that according to the specific topology of pedestrians, different horizontal stripes correspond to different semantic regions of human bodies. Thus, we can align semantic-group textual features to different local visual features according to semantic cues, which achieves implicit fine-grained image-text feature alignment.
To guide semantic-group textual features aligning with visual concepts, we 
design a Vision-Guided Knowledge Transfer (VGKT) module, including a vision-guided attention and a relational knowledge transfer.
The former exploits visual concepts to extract vision-guided textual features that contain relevant textual features according to their paired visual features. In this way, textual features contain semantic cues consistent with visual concepts and have high correspondence with visual concepts.
However, the paired visual features can only be obtained during the training process due to the need for identity labels. 
Inspired by knowledge distillation of Hinton~\cite{hinton2015distilling}, 
we aim to transfer the information learned from vision-language interaction into the paradigm of self-learning. 
\ie,
we transfer the relational knowledge from vision-guided textual features as teacher to semantic-group textual features learning as shown in \figurename~\ref{fig:motivation}(d). Relation constraints consisting of a vision-language similarity and a class probability naturally fit the objective function of text-based person search task. 
The proposed VGKT facilitates aligning semantic-group textual features with local visual features implicitly and efficiently without complex cross-modal interaction.

Our main contributions are summarised as follows:

\begin{itemize}
\setlength\itemsep{0.5em}
    \item We propose a Vision-Guided Semantic-Group Network (\ourmodel) to alleviate the misalignment of fine-grained cross-modal features by semantically grouping textual features and then aligning them with visual concepts.

    \item We build a strong CLIP-driven baseline framework to facilitate aligning fine-grained cross-modal features.
    \item We develop a Semantic-Group Textual Learning (SGTL) module to group semantic patterns in  channel dimension under the global context.
    \item We design a Vision-Guided Knowledge Transfer (VGKT) module to align semantic-group textual features with visual concepts without the assistance of semantic annotations.
    \item We achieve new state-of-the-art performance on two text-based person search benchmarks, CUHK-PEDES~\cite{li2017person} and ICFG-PEDES~\cite{ssan}.

\end{itemize}

\section{Related Work}

\subsection{Text-based Person Search}
Text-based person search aims to match images of the target person with free-form natural languages. Li~\etal~\cite{li2017person} build a challenging datasets CUHK-PEDES with detailed language annotations and propose text-based person search for the first time.
There are two mainstreams in text-based person search:
pairwise similarity learning~\cite{li2017person, li2017identity, chen2018improving} and joint embedding learning~\cite{zheng2020dual, zhang2018cmpc,he2023region,wang2020vitaa}.

Li~\etal~\cite{li2017person} introduce a recurrent neural network with gated neural attention (GNA-RNN) to learning similarity between sentence and a person image.
Both the unit-level attention (for individual visual units) and word-level sigmoid gates  
are utilized to build cross-modal relation.
Li~\etal~\cite{li2017identity} further propose a two-stage matching network which exploits
identity-level annotations to learn discriminative feature representations. Two stages are complementary to each other and both contribute to get robust matching result.
Chen~\etal~\cite{chen2018improving} design patch-word matching framework to capture
the local matching details across modalities. In addition, a adaptive threshold is introduced to alleviate its susceptibility to the matching degree of corresponding image-word pair.
Chen~\etal~\cite{R11} propose a simple, convenient, and real-time person search system to learn a joint latent space among audio, text, and visual modalities in a progressive way with great practical value.

While these methods are time-consuming and more and more researchers alter their attention to joint embedding learning which calculate the affinity for image-text pairs in a shared feature space.
Zheng~\etal~\cite{zheng2020dual} propose instance loss to take intra-modal distribution into account explicitly which contribute to learning the discriminative representation from every image/text group.
Zhang~\etal~\cite{zhang2018cmpc} propose a cross-modal projection matching (CMPM) loss and a cross-modal projection classification (CMPC) to enhance inter-class distinctiveness and intra-class compactness. which advances capturing discriminative image-text embeddings.

The above approaches are computation efficient, but they ignore part or fine-grained feature extraction which are critical for text-base person search. Recently, there are some attempts at extracting multi-granularity fine-grained representations~\cite{ssan,jing2020pose,wang2021mgel, farooq2021axm,zheng2020hierarchical,chen2021tipcb}.
Jing~\etal~\cite{jing2020pose} utilize human pose network to align cross-modal fine-grained features.
Ding~\etal~\cite{ssan} introduce a semantically self-aligned network (SSAN) utilizing attention mechanisms to extract text features corresponding to part-level visual features.
Farooq~\etal~\cite{farooq2021axm} propose AXM-Net, which dynamically leverages multi-scale knowledge from both modalities and recalibrates each modality based on shared semantics.
Although these methods enhance local information and demonstrate improvements, they either rely on external tools such as pose alignment or focus solely on language information, overlooking image information in guiding language. In contrast, our approach extracts corresponding text descriptions for part-level features via semantic grouping and achieves cross-modal alignment by distilling knowledge from vision, without the need for extra tools or annotations.

\subsection{Visual-Language Pretraining}
Visual-Language pretraining has attracted more and more attention nowadays and achieved remarkable performance on different multi-modal downstream tasks. Several methods~\cite{clip, milnce,simvlm} exploit large-scale data source and utilize semantic supervision to learn visual representations from text representations. 
MIL-NCE~\cite{milnce} has ability of addressing misalignments problem in narrated videos to learn strong video representations rom noisy large-scale HowTo100M~\cite{howto100m}.
SimVLM~\cite{simvlm} reduces the training complexity by exploiting large-scale weak supervision, and is trained end-to-end with a single prefix language modeling objective.
Recently, Contrastive Language-Image Pretraining (CLIP)~\cite{clip} has achieved superior results in multi-modal zero-shot learning, which shows image and text representations can be well aligned in the embedding space. Its remarkable generalization capability is benefit from large-scale training samples (400 million text-image pairs collected from Internet).
It has been transferred to different downstream tasks and shown promising results, such as image generation~\cite{hairclip}, zero-shot video retrieval~\cite{clip2video}, video caption~\cite{clip4caption} and zero-shot segmentation~\cite{PADing,3DZero,D2Zero}.
TextReID~\cite{textreid} utilize CLIP to match image-text pair in text-based person search for the first time.
But it exploits a redundant Bi-GRU~\cite{cho2014gru} to construct cross-modal alignment between image and text, which is very complicated and inefficient as it ignores the strong cross-modal alignment capability inherent in CLIP.
Our work is built upon the pure CLIP and leverages its multi-modal alignment capability for image-text matching. In addition, we take into account the fine-grained information inherent from local visual and textual features neglected by CLIP.

\subsection{Attention and Transformer}

The Transformer is proposed by vaswani~\cite{vaswani2017transformer} to handle sequential data using only the attention mechanisms. 
It has become the de-facto standard for natural
language processing (NLP) tasks such as machine translation~\cite{vaswani2017transformer}, speech processing~\cite{ren2019fastspeech}, question answering~\cite{devlin2018bert}. 
Inspired by NLP successes, various works adopt the transformer in the computer vision field~\cite{VLTTPAMI,GRES,MeViS} and shows promising performance on multiple tasks such as object detection~\cite{detr}, semantic segmentation~\cite{strudel2021segmenter, ding2018context}, image recognition~\cite{vit}, object re-identification or tracking~\cite{transreid,MOSE}, etc. 
The merit inherent in transformer is its capability of exploring long-range dependency.

Several efforts~\cite{ssan,jing2020pose,wang2021mgel,farooq2021axm,zheng2020hierarchical,gao2021contextual} have been made to introduce the attention mechanism in the text-based person search. 
Gao~\etal~\cite{gao2021contextual} put forward a contextual non-local attention mechanism to automatically align multi-scale visual features with textual features.
However, this paradigm requires repetitive feature alignment for each image-text pair through the cross-modal attention module, resulting in considerable computational overhead during inference. In our method, we propose a vision-guided attention strategy that obtains vision-guided textual features as a teacher, achieving implicit and efficient alignment between local textual and visual features without intricate pairwise cross-modal interactions.

\subsection{Knowledge Distillation}
Knowledge distillation~\cite{hinton2015distilling} transfer knowledge from a large teacher model to a small student model so as to improve the ability of student model.
Through imitating teacher class probabilities or feature representations, knowledge distillation provides additional supervision that differs from conventional supervised learning objectives~\cite{parisotto2015actor,romero2014fitnets,chen2018darkrank}.
Chen~\etal~\cite{R12} introduce a metric distillation loss that enables the interpreter to disentangle the distance between two individuals into attribute components, leveraging knowledge distilled from the target model.
Zhang~\etal~\cite{R21} incorporate cross-modal knowledge transfer from visual to audio domains within a semi-supervised learning framework, employing consistency regularization to enhance pseudo-label quality, thereby augmenting model diversity and robustness.
Nonetheless, these approaches are not specifically designed for part-level image-text alignment in the context of ReID. Building on these foundations, we propose a relational knowledge transfer mechanism that guides semantic-group textual features by distilling intrinsic relational knowledge from vision-guided textual features. Our relational knowledge transfer includes vision-language similarity and class probability transfer, seamlessly aligning with the objective function of text-based person search tasks.

\begin{figure*}[t]
	\centering
	\includegraphics[width = \textwidth]{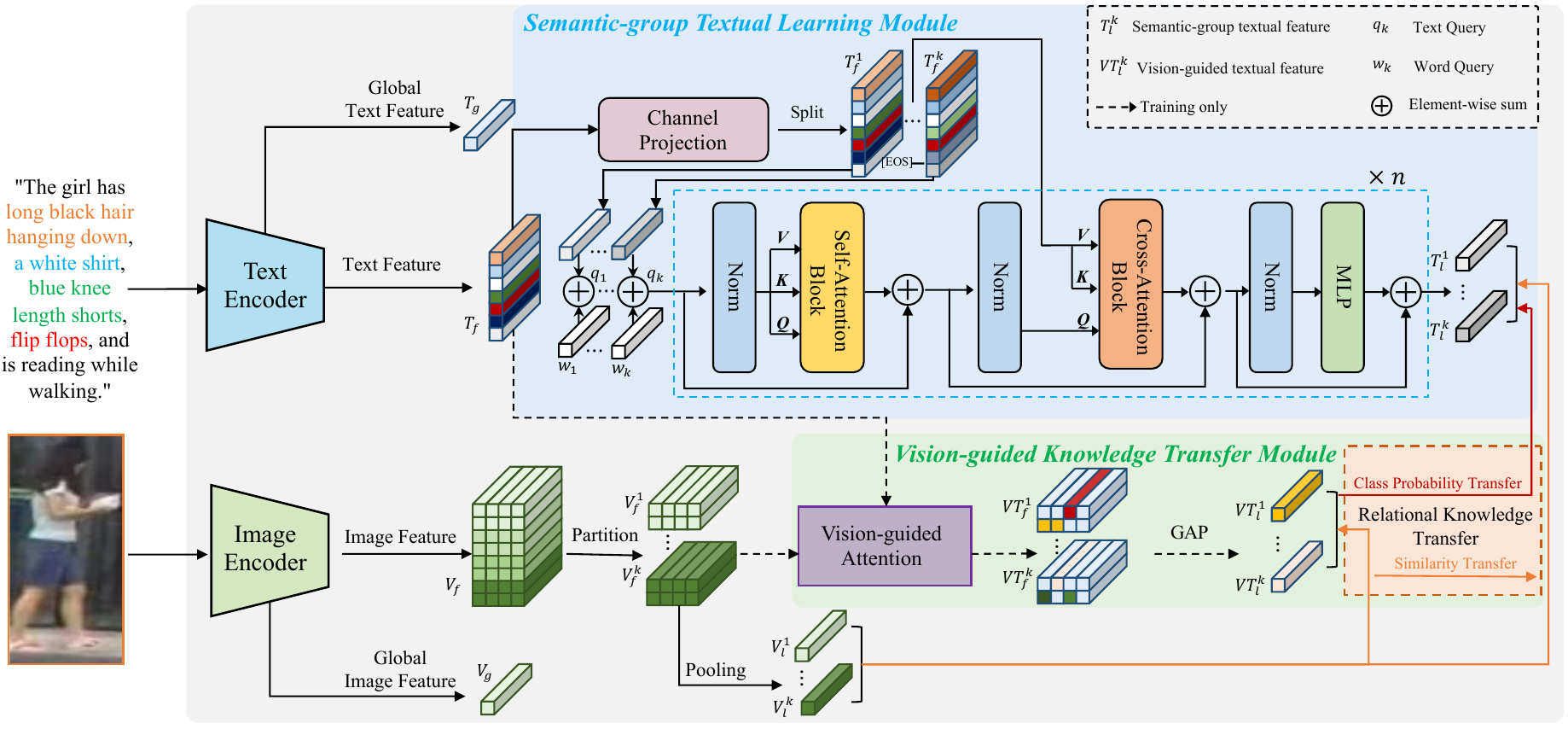}
        \vspace{-4mm}
	\caption{\small{The architecture overview of the proposed \ourmodel. We embed image and text into global image and text features by an image encoder and a text encoder. Semantic-Group Textual Learning (SGTL) module learns semantic-group local textual features from the textual feature by applying channel-group and attention mechanism. Vision-Guided Knowledge Transfer (VGKT) module consists of Vision-guided Attention and Relational Knowledge Transfer. The former extracts the relevant textual information for each local visual features, and the latter propagates the textual information flow to align the semantic-group textual features with local visual features. Both global feature and local features are utilized to calculate loss function.}}
	\label{fig:framework}
\end{figure*}

\section{Approach}

\subsection{Revisiting CLIP}
CLIP~\cite{clip} is a superior vision-language pre-training model that achieves impressive performance in associating visual and textual concepts. It consists of two independent encoders: an image encoder $E_v$ and a text encoder $E_t$, respectively. For a given image-text pair $\{I, T\}$, CLIP first extracts image feature $V_f = E_v(I)$ and text features $T_f = E_t(T)$ and then calculate cosine similarity $\mathcal{S} = \langle V_f, T_f\rangle$ between these two features. A contrastive loss is imposed on $\mathcal{S}$ to force image features and their corresponding text features to be close to each other, which facilitates image features and their paired text features being aligned accordingly. However, CLIP trained on image classification image-text pairs computes global similarity of visual and textual features while ignoring the fine-grained details of text and image, and thus cannot be directly used in text-based person search that requires local fine-grained clues.

\subsection{CLIP-driven Baseline for Text-based Person Search}\label{sec:strong-baseline}

As shown in \figurename~\ref{fig:framework}, our model takes an image $I$ containing a person, and a sentence $T$ describing the attributes of the person as inputs. We utilize ResNet-50~\cite{he2016resnet} as our image encoder to extract visual feature $V_{f}\in \mathbb{R}^{C\times H \times W}$ from image $I$, where $C$, $H$, and $W$ denote the number of channels, height and width of the feature maps, respectively. Global average pooling is replaced with self-attention pooling layer over feature map $V_{f}$ to get global image feature $V_{g}\in \mathbb{R}^{C}$. For an input sentence $T\in \mathbb{R}^{L}$, where $L$ denote the length of a given sentence, we utilize Transformer~\cite{vaswani2017transformer} as our text encoder to extract text feature $T_{f}\in \mathbb{R}^{L\times C^T}$, where $C^T$ is the feature dimension of each word. Words of the sentence are encoded according to byte pair encoding (BPE) with a 49,152 vocab size. \texttt{[SOS]} and \texttt{[EOS]} tokens are prepended and postpended to text sequence tokens, indicating the beginning and the end of a sentence, respectively. The representation of the last layer of the transformer is the global text representation, after which a text projection is applied to get textual feature $T_{g}\in \mathbb{R}^{C}$ that has the same channel number as the $V_{g}$.

The similarity between global image feature and text feature is obtained as follows:
\begin{equation}
\mathcal{S}_g = \langle V_g, T_g\rangle = \frac{V_g^T T_g}{||V_g||||T_g||},
\end{equation}
where $\langle,\rangle$ denotes cosine similarity. We utilize ID loss and contrastive loss on global feature to optimize the network. The ID loss $\mathcal{L}_{ID}$ is the cross-entropy loss with label smoothing and imposed both on $V_g$ and $T_g$. 
For a triplet set $\{V_{g,i}^{+}, T_{g,i}^{+}, T_{g,i}^{-}\}$, where $T_{g,i}^{+}$ and $T_{g,i}^{-}$ are the positive pair and negative pair of $V_{g,i}^{+}$, receptively, we adopt contrastive loss \cite{wang2020vitaa} as follows: 
\begin{equation}
\label{align_loss}
\mathcal{L}_{Con}(g)=\frac{1}{B} \sum_{i=1}^{B}\left\{\begin{aligned}\log \left[1+e^{-\tau_{p}\left(\mathcal{S}_{g,i}^{+}-\alpha\right)}\right]\\+\log \left[1+e^{\tau_{n}\left(\mathcal{S}_{g,i}^{-}-\beta\right)}\right]\end{aligned}\right\},
\end{equation}
where $\mathcal{S}_{g,i}^{+} = \langle V_{g,i}^{+}, T_{g,i}^{+}\rangle$, $\mathcal{S}_{g,i}^{-} = \langle V_{g,i}^{+}, T_{g,i}^{-}\rangle$. $\tau_{p}/\tau_{n}$ and $\alpha/\beta$ denotes the temperature parameters and absolute margin bound for positive/negative pairs, respectively. $B$ is the training batch size. We set $\tau_{p}/\tau_{n}$ and $\alpha/\beta$ to 10/40 and 0.6/0.4 in the all experiments. With Eq.~(\ref{align_loss}), we force image global feature of the target person to be close with its corresponding text feature.

\subsection{Semantic-Group Textual Learning Module}\label{sec:sgtl}
Although CLIP-driven baseline has achieved impressive performance in text-based person search, it mainly focuses global information from the entire image and the whole sentence. However, learning fine-grained local features is critical for fine-grained cross-modal retrieval, especially for person search~\cite{ssan,jing2020pose,wang2021mgel,farooq2021axm,zheng2020hierarchical,chen2021tipcb} that distinguishes each identity with detail information. 
Here, we aim to learn local visual and textual features to extract discriminative local cues.
For local visual feature extraction, we follow previous methods~\cite{ssan,chen2021tipcb,zheng2020hierarchical,farooq2021axm,wang2021mgel} and uniformly partition visual feature $V_f$ into $K$ non-overlapping local image feature maps, denoted as $\{V_{f}^k\}_{k=1}^K$.
After that, self-attention pooling layer, which is not shared with global branch, is applied on $\{V_{f}^k\}_{k=1}^K$ to get local-pooling image feature $\{V_{l}^k\}_{k=1}^K$.
But for fine-grained language information extraction, it cannot be horizontally split like visual features for lack of spatial information.
Existing methods~\cite{farooq2021axm,jing2020pose,zheng2020hierarchical} utilize external tools like NLTK~\cite{loper2002nltk} to detect noun phrases as local discriminative cues. However, this partition method is too rough, and does not take the contextual textual information into account. Moreover, the final performance depends on the quality of external tools largely. {Herein, we propose to extract local representations of textual features via grouping semantic features without using any external tools.}

Specifically, text feature $T_f$ is fed into Channel Projection block (see \figurename~\ref{fig:framework}), which is composed of linear layer followed by batch normalization layer, to expand channel dimension from $C^T$ to $KC^{T}$.   
After that, we split it into $K$ groups from the channel dimension to obtain a new feature set $\{T_f^k\}_{k=1}^K \in \mathbb{R}^{L\times C^T}$. 
Each channel-group textual feature $T_f^k$ is expected to be related to its corresponding local visual feature $V_f^k$ according to semantic distribution, under the supervision presented in Sec.~\ref{sec:vgkt}.

Then, to detect important phrases, we introduce a set of learnable word queries $\{w^k\}_{k=1}^K$ and $w^k \in \mathbb{R}^{C^T}$.
Apart from word queries, we add conditional global textual representation $T_{f\texttt{[EOS]}}^k\in \mathbb{R}^{C^T}$ for better convergence as well as fully utilizing specific textual information.
Hence, the input of Transformer block is calculated by $q^k = w^k + T_{f\texttt{[EOS]}}^k$ and $q^k$ is text query. 
The transformer block is composed of $n$ stacked layers to excavate semantic cues from textual features. 
Following the standard architecture of the transformer \cite{vaswani2017transformer},
we first apply multi-head self-attention layer to propagate information flow between text query including textual features and word query:
\begin{equation}
    q^{k\prime} = MHA(LN(q^k), LN(q^k), LN(q^k)) + q^k,
    \label{eq5}
\end{equation}
where $q^{k\prime}$ is the evolved text query feature, $LN(\cdot)$ and $MHA(\cdot)$ are layer normalization and multi-head attention, respectively.
Multi-head attention is performed as follows,
\begin{equation}
    MHA(Q,\, K,\, V) = softmax(\frac{QK^{T}}{\sqrt{d_k}})V.
    \label{eq6}
\end{equation}
The multi-head attention layer consists of three independent linear layers mapping inputs including Q, K, and V to the same intermediate representations of dimension $d_k$. 
Q, K, and V are the same for multi-head self-attention in Eq.~(\ref{eq5}).
Subsequently, multi-head cross-attention layer is utilized to
enable similar semantic patterns to be grouped together to obtain semantic-group local textual features with text query. The queries come from text query, keys and values come from local textual feature. 
The output is further processed by an MLP block of two fully-connected layers accompanied by layer normalization and residual connections:
\begin{equation}
\begin{aligned}
    &T_{l}^{k\prime} = MHA(LN(q^{k\prime}),\,T_f^k,\,T_f^k) + q^{k\prime},\\
    &T_{l}^k = MLP(LN(T_{l}^{k\prime})) + T_{l}^{k\prime},
    \label{eq7}
\end{aligned}
\end{equation}
where $T_{l}^{k\prime}$ is the intermediate features and $T_{l}^k$ is the final extracted semantic-group local textual feature. 

So far, we obtain local visual features and their corresponding semantic-group local textual features.
The similarity $\mathcal{S}_l$ between them is calculated as follows:
\begin{equation}
\mathcal{S}_l = \langle V_l, T_l\rangle=\frac{V_l^T T_l}{||V_l||||T_l||},
\label{eq:score}
\end{equation}
where $V_l$ and $T_l \in \mathbb{R}^{KC}$ are calculated by concatenating the $K$ local visual and textual features, respectively.

\subsection{Vision-guided Knowledge Transfer Module}\label{sec:vgkt}
The semantic-group textual features from SGTL are learned from the textual feature itself, which does not pay attention to the distribution of the local visual feature. As such,
semantic-group textual features cannot capture accurate semantic patterns aligned
with corresponding visual concepts without the assistance of
semantic annotations. Therefore, to guide semantic-group
textual features aligning with visual concepts, we design a
Vision-Guided Knowledge Transfer (VGKT) module including Vision-guided Attention and Relational Knowledge Transfer. 

\begin{figure}[t]
	\centering
	\includegraphics[width = 0.48\textwidth]{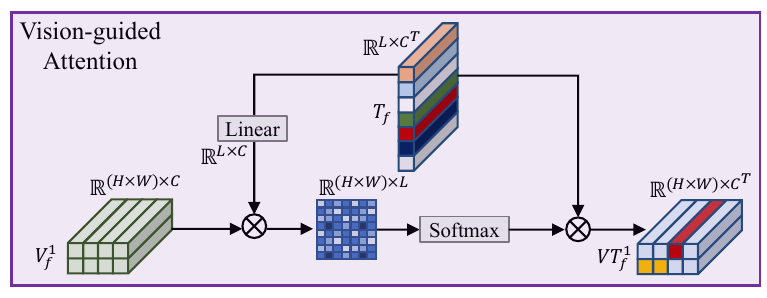}
	\caption{\small{The architecture of our proposed Vision-guided Attention module. It has capable of extracting relevant textual features aligned with visual features.}}
	\label{fig:vga}
\end{figure}

\paragraph{Vision-guided Attention}
Vision-guided attention aims to extract local text representations corresponding to local visual feature. 
As shown in \figurename~\ref{fig:vga},
First, we compare each word feature with the visual region features to determine the importance of each word feature. For the $i$-th pixel of the image feature map, we calculate its attention weights focusing on the $j$-th word as:
\begin{equation}
\alpha_{i,j} = \frac{exp(V_{f,i}^{T} W_1 T_{f,j})}{\sum_{j=1}^L exp(V_{f,i}^{T} W_1 T_{f,j})},
\end{equation}
where $W_1\in \mathbb{R}^{C^T\times C}$ is the learnable parameter of a fully connected layer. 
If $i$-th pixel has an obvious correlation with word $T_{f,j}$, it will output a high attention score $\alpha_{i,j}$. Otherwise, it will output a low attention score for the $j$-th word. 
A set of attention weights $\{\alpha_{i,j}\}_{i=1}^{H\times W},_{j=1}^L$ is hence generated.
Then, we extract the features for each pixel of the image based on the attention weights. For example, we get the $i$-th feature $VT_{f,i}$ by weighted aggregating textual features of all words:
\begin{equation}
VT_{f,i} = \sum_{j=1}^L \alpha_{i,j} T_{f,j}.
\end{equation}
Intuitively, $VT_{f,i}$ captures relevant textual evidence according to visual information of $i-$th pixel.
Thus, we can get a set of feature
$VT_f = \{VT_{f,1},VT_{f,2},\dots, VT_{f, H\times W}\}$ and $\{VT_{f}^k\}_{k=1}^K$ according to the visual feature partition. Finally, we utilize global average pooling on $\{VT_{f}^k\}_{k=1}^K$ to get vision-guided textual feature $\{VT_{l}^k\}_{k=1}^K$. In general, for each local visual feature,  
we utilize the relationship between visual and language to extract relevant text features correlated with visual features.
The similarity $\mathcal{S}_l^v$ between local visual feature and vision-guided textual feature is calculated as follows:
\begin{equation}
\mathcal{S}_l^v = \langle V_l, VT_l\rangle = \frac{V_l^T VT_l}{||V_l||||VT_l||},
\label{eq:score2}
\end{equation}
where $V_l$ and $VT_l \in \mathbb{R}^{KC}$ are calculated by concatenating the $K$ local visual and textual features, respectively.

\paragraph{Relational Knowledge Transfer}
The paired visual and textual features can be only obtained during the training process due to the need for identity labels. During the inference stage, the corresponding vision-guided text feature can not be obtained unless in the form of pairwise-learning, which is time-consuming and has very high computational complexity. Inspired by knowledge distillation~\cite{hinton2015distilling},
we propose to transfer the information learned from vision-language interaction into the paradigm of self-learning. 
Specifically, we leverage the information of feature $VT_l^{k}$ as supervision signals to constrain the feature $T_l^{k}$ generation. When there is no visual feature assistance, the textual feature $T_l^{k}$ still has the capability to learn the corresponding visual local information following the distribution of teacher feature $VT_l^{k}$.

\textit{Vision-Language Similarity Transfer:}
Inspire by \cite{cao2007learning,xia2008listwise}, we utilize the pairwise similarity between two modalities (visual and textual) as complementary supervision, instead of concentrating on the feature representation alone. In the retrieval task, the similarity between features directly determines the final performance so we propose to transfer this dark relationship knowledge. Here, we denote relation matrix $M=\{m_{ij}\}_{i=1}^B,_{j=1}^B\in \mathbb{R}^{B \times B}$ and relation value $m_{ij}$ is computed by similarity $\mathcal{S}_{ij}$ between visual and textual features and further normalized by softmax function:
\begin{equation}
\label{eqn:permprob}
m_{ij}= \frac{\exp(\mathcal{S}_{ij}/\tau)}{\sum_{\hat{j}=1}^{B}\exp(\mathcal{S}_{i\hat{j}}/\tau)},
\end{equation}
where $\tau$ is a temperature factor to sharpen the relation distribution. 
Thus, depending on the different types of local textual features, we have the vision-guided textual feature relation matrix as $M^v$. For the semantic-group textual feature, we have the relation matrix as $M^s$.

Then, we transfer the similarity relation from vision-guided textual feature to semantic-group textual feature with KL divergence.
Vision-language similarity distillation loss is defined as 
\begin{equation}
\mathcal{L}_{st}(M^s, M^v) = D_\text{KL}[M^v~\| M^s)],
\end{equation} 
where $D_{KL}[p||q]=-\int p(z) log {p(z) \over q(z)}$. The vision-language similarity loss transfers the relationship of vision $\rightarrow$ vision-guided textual feature to that of vision $\rightarrow$ semantic-group textual feature, which fits the objective of the retrieval task. It is worth noting that $\mathcal{L}_{st}$ won't be back-propagated through the vision-guided attention module during model training. In that way, the vision-guided attention remains the capability to extract vision-guided textual features without being interfered by other loss functions.

\textit{Class Probability Transfer:}
Apart from vision-language similarity transfer, class probability~\cite{chen2021cmka} is essential to the discriminative capability of the model. 
Analogously, we design class probability transfer to further adapt predicted class probability from the vision-guided textual feature to the class probability produced by semantic-group textual feature. Formally, we define class probability matrix $P^k=\{p_{ij}^k\}_{i=1}^B,_{j=1}^N\in \mathbb{R}^{B \times N}$, where $p_{ij}^k$ is class probability value calculated by passing the $k$-th local textual features through the classifier with softmax function.
Thus, for vision-guided textual feature, we have the class probability matrix as $P^{v,k}$. In the semantic-group textual feature, we have the class probability matrix as $P^{s,k}$.
Then, we transfer the class probability relation from vision-guided textual feature to semantic-group textual feature with KL divergence. Class probability distillation loss is defined as:
\begin{equation}
\mathcal{L}_{cpt}(P^{s,k}, P^{v,k}) = \sum_{k=1}^K D_\text{KL}[P^{v,k}~\| P^{s,k})].
\end{equation} 
$\mathcal{L}_{cpt}$ is not back-propagated through the vision-guided module during model training too.

These two losses, $\mathcal{L}_{st}$ and $\mathcal{L}_{cpt}$, aim to migrate the relation knowledge from the vision-guided textual feature to semantic-group textual feature, which facilitates aligning semantic-group textual feature with local visual features. 

\subsection{Training and Inference}
During training, $\mathcal{L}_{ID}$ is imposed on $V_g$, $T_g$, $K$ local visual features, $K$ semantic-group local textual features and $K$ vision-guided local textual features.
$\mathcal{L}_{Con}$ is imposed on global image-text pair $\mathcal{S}_g$, semantic-group local image-text pair $\mathcal{S}_l$ and vision-guided local image-text pair $\mathcal{S}_l^v$. The overall loss is computed as follows:
\begin{equation}
\mathcal{L}=\mathcal{L}_{ID} + \mathcal{L}_{Con} +\lambda_1\mathcal{L}_{st}+\lambda_2\mathcal{L}_{cpt},
\label{eq:loss_function}
\end{equation}
where $\lambda_1$, $\lambda_2$ are the weights to balance the effects of different losses.
During the inference stage, the overall similarity value is the sum of global image-text pair $\mathcal{S}_g$ and semantic-group local image-text pair $\mathcal{S}_l$.

\section{Experiments}

\subsection{Datasets}
\textbf{CUHK-PEDES}~\cite{li2017person} is made up of 40,206 images and 80,412 language descriptions for 13,003 identities where each image include two sentences. The training set consists of 34,054 images and 68,108 language descriptions for 11,003 identities. Both validation set and test set have 1,000 identities where 3,078 images are in the validation set and 3,074 images are in the test set. The language description comprises 23 words on average.

\textbf{ICFG-PEDES}~\cite{ssan} is constructed from the MSMT17~\cite{MSMT17} database which is made up of 54,522 images and 54,522 language descriptions for 4,102 identities where each image include only one sentences.
The training set consists of 34,674 images for 3,102 identities. The test set has 1,000 identities with 19,848 images.
The language description comprises 37 words on average.

\renewcommand{\multirowsetup}{\centering}
\begin{table*}[t]
\renewcommand\arraystretch{1.06}
    \caption{ \small{The component analysis of \ourmodel. Local means adding $K$ convolution layer to extract local textual feature. SGTL and VGKT are our proposed Semantic-Group Textual Learning Module and Vision-guided Knowledge Transfer Module, respectively.}}\label{tab:ablation}
    \vspace{-2mm}
\footnotesize
    \begin{center}
    \setlength{\tabcolsep}{3.96mm}{\begin{tabular}{lccc|ccc|ccc}
    \hline
     & & & &\multicolumn{3}{c|}{CUHK-PEDES} & \multicolumn{3}{c}{ICFG-PEDES} \\
    Index   & Local & SGTL & VGKT   & Rank-1    & Rank-5 & Rank-10    & Rank-1 & Rank-5 & Rank-10  \\
    \hline
    \hline
    1  &\xmarkg &  \xmarkg&  \xmarkg     &  65.26  &  82.81  &   88.84  & 56.69 &72.96&79.45\\
    2& \cmark & \xmarkg &      \xmarkg      & 65.58 & 83.05  &  89.17 & 56.90&72.58&  78.91 \\
    3& \xmarkg& \cmark     &\xmarkg  & 66.48  & 83.64  &  89.52 & 58.85& 74.41& 80.40 \\
    4& \cmark &  \xmarkg    & \cmark     &  66.76  & 83.56  &  89.32 &   59.13        &     74.66      & 80.73\\
    5~(\textbf{\ourmodel})& \xmarkg &  \cmark    & \cmark     &  67.52  & 84.37  &  90.26 &   60.34        &     76.01      & 82.01\\
    \hline
    \end{tabular}}
\end{center}

\end{table*}
\subsection{Implementation}

\textbf{Experimental Details:}
We utilize the image and text encoder with CLIP~\cite{clip}, and choose ResNet-50\cite{he2016resnet} as the image encoder for all the experiments. CLIP-ViT visual backbone is applied for fair SOTA performance comparison.
The maximum length of input sentences is set to 77.
All person images are resized to $256 \times 128$ unless otherwise specified. 
Following conventions in the ReID community, the stride of the last block in the ResNet is set to 1 to increase the resolution of the final feature map.
The training images are augmented with random horizontal flipping, padding, random cropping, and random erasing \cite{random_erase3}. 
The batch size is set to 96 with 4 images and sentences pairs per identity. 
We train the network for 50 epochs employing the Adam optimizer with the initial learning rate of 0.00035, which is decayed by a factor of 0.1 after the 40th epoch. 
Temperature $\tau$ and the transformer layer $n$ are set to 4 and 2, respectively. 
All the experiments are performed with one RTX TITAN using the PyTorch toolbox.
\footnote{http://pytorch.org}. 

\textbf{Evaluation Protocols.} Following previous work~\cite{textreid}, we evaluate all methods with Rank-K(K=1,5,10) and some with the mean Average Precision (mAP). 
Rank-K indicates the percentage that, given a text/image as query, more than one correct image/text are retrieved among the top-k candidate list.
mAP is adopted for a comprehensive evaluation that focuses more on the order of the total retrieval results.

\subsection{Ablation Study}

\paragraph{Component Analysis}

We perform detailed component ablation studies to evaluate the effectiveness of our proposed method in \tablename~\ref{tab:ablation}. The experiments are performed on both CUHK-PEDES and ICFG-PEDES. 
We adopt the method clarified in Sec.~\ref{sec:strong-baseline} as our baseline with global feature only (index 1). 
First of all, when adding the proposed SGTL into our baseline (index 3), a performance gain of 1.22\% and 2.16\% in terms of Rank-1 is observed over the baseline on CUHK-PEDES and ICFG-PEDES, respectively. The performance gain is owning to the benefit of our effective local textual learning design.
The superior result demonstrates that  similar semantic patterns are grouped
together to represent semantic-group local textual features, which is important in text-based person search requiring fine-grained information.
On the other hand, to verify the superiority of our method, we use Local which adds $K$ convolution layer to extract local textual features instead of SGTL (index 2). A performance gain is small with 0.32\% and 0.21\% in terms of Rank-1,
which proves that the gain of SGTL is not from increasing the number of parameters or introducing additional local visual features.

Then, by introducing VGKT as auxiliary supervision (index 5), we further get a significant improvement over the SGTL, \eg, 1.04\% and 1.49\% performance gain in terms of Rank-1 on CUHK-PEDES and ICFG-PEDES, respectively. 
The proposed VGKT forces the knowledge to adapt from vision-guided textual features to semantic-group textual features. With the help of VGKT,
SGTL generation is no longer only depending on itself, but taking its paired visual features into account through knowledge distillation loss in VGKT. As such, semantic-group textual features are capable of aligning local visual features.
To further illustrate its effectiveness, We add VGKT built upon Local (index 4).
There is also a significant performance gain beyond Local, which verifies that our VGKT is a versatile module.


\paragraph{Ablation Study of Baseline Configuration}
We study the effects of various designs of the baseline configuration. 
The results of different variations of the training settings are listed in \tablename~\ref{tab:ablation-baseline}. 
The results are improved consistently with the help of opening text encoder, adding drop path, and building shared layer with visual and textual embedding.
Because text-based person search is a fine-grained task that tackles in the same category. If we freeze the text encoder like other methods~\cite{textreid}, the differences in text embedding are very limited. Therefore, we finetune the text encoder in the TBPS dataset with a small learning rate of 3e-6 to capture fine-grained information. When freezing the text encoder, the performance decreases by 14.39\% Rank-1 and 10.2\% Rank-1 on CUHK-PEDES and ICFG-PEDES, respectively.
Introducing stochastic depth~\cite{stoc_depth} can boost the Rank-1 performance by about 2.61\% and 2.41\% on CUHK-PEDES and ICFG-PEDES, respectively. It is because the transformer design in text encoder has no regularization components, it is easily overfitting in downstream tasks without large datasets.
Constructing share layer between visual and textual modality can reduce the modality gap~\cite{zheng2020dual,ssan}, which provides 0.25\% Rank-1 improvement on ICFG-PEDES compared with the not share layer.

\begin{table}[t]
\renewcommand\arraystretch{1.06}
	\centering
		\caption{\small{Effects of different baseline network configuration on CUHK-PEDES and ICFG-PEDES datasets. The abbreviations O, D, S denote opening text encoder, stochastic depth~\cite{stoc_depth} and share layer between visual and textual modality, respectively.}}
    \setlength{\tabcolsep}{3.36mm}{\begin{tabular}{ccc|cc|cc}
    \hline
    			 &  &  & \multicolumn{2}{c|}{CUHK-PEDES} & \multicolumn{2}{c}{ICFG-PEDES} \\  
			O & D & S & Rank-1 & Rank-5 &Rank-1  & Rank-5 \\ \hline  \hline 
		    \xmarkg  & \xmarkg  & \xmarkg &   48.26 &  70.31  &   39.04 &   57.82  \\
			\cmark  & \xmarkg  & \xmarkg  &  62.65  &  80.40  &   53.93 &   70.83  \\
			\cmark  & \cmark  & \xmarkg   &  65.06  &  81.43  &   56.34 &   72.12  \\
			\cmark  & \cmark  & \cmark    &   65.26  &  82.81  &   56.59 &   72.96\\
	\hline
	\end{tabular}}

	\label{tab:ablation-baseline}
\end{table}
\subsection{Ablation Study of SGTL}

\paragraph{Evaluation of the Number of $K$}
We conduct several experiments to investigate the influence of the different number of local visual and textual features in~\figurename~\ref{fig:hyper-parameter} (a). As can be seen, with $K$ increase, the performance first improves and then becomes flat. It strikes the peak when $K$ arrives 4, the trend is similar to the past methods like PCB~\cite{sun2018beyond} and TransReID~\cite{transreid}. While overly large region numbers consume more computing resources and slow down the inference speed. Thus, we choose 4 to achieve a good balance between accuracy and efficiency. 

\begin{table}[t]
\renewcommand\arraystretch{1.06}
	\centering
	\caption{\small{Comparison with other designs for SGTL on CUHK-PEDES and ICFG-PEDES datasets. TQ and CG denote text query and channel group, respectively.}}
    \setlength{\tabcolsep}{2.76mm}{\begin{tabular}{ccc|cc|cc}
    \hline
    	\multirow{2}{*}{Index}& 		\multirow{2}{*}{TQ}& \multirow{2}{*}{CG} & \multicolumn{2}{c|}{CUHK-PEDES} & \multicolumn{2}{c}{ICFG-PEDES} \\  
		&	  &  & Rank-1 & Rank-5 &Rank-1  & Rank-5 \\ \hline  \hline 
		1 &    \xmarkg  & \xmarkg &  65.73 &  83.13  &   57.15  &   72.96  \\
		2&	 \cmark  & \xmarkg  &  66.01  &  83.21  &   57.71  &   73.43  \\
		3&	 \xmarkg  & \cmark   &  66.29  & 83.26  & 57.66   &  73.72   \\
		4&	 \cmark  & \cmark    &   66.48 & 83.64  &   58.85  &  74.41\\
	\hline
	\end{tabular}}
	\label{tab:SGTL-variations}
\end{table}
\paragraph{Comparison with Other Designs for SGTL}

To verify the superiority of our proposed SGTL, we conduct experiments on other variations for SGTL in \tablename~\ref{tab:SGTL-variations}.
For a fair comparison, all the methods are performed with the same experimental configuration apart from the local textual learning design.
on the one hand,
We use DETR~\cite{detr} decoder branch as variants that contains self-and cross-attention block design without considering channel group operation. We set the text query as \textit{querys}, the whole textual features $T_f$ as the \textit{keys} and \textit{values}. Through transformer learning, text queries aim to attend to different local discriminative cues from the entire textual features.
As the experimental results shown in \tablename~\ref{tab:SGTL-variations} (index 2), vanilla DETR transformer performs inferior to our proposed SGTL (index 4), where performance drops of -0.47\% on CUHK-PEDES and -1.14\% on ICFG-PEDES in terms of Rank-1 accuracy are observed, respectively. This confirms that simply applying Transformers for the text-based person search is not effective, because they neglect that feature channels
of textual feature can be grouped according to semantic
distribution and corresponds to a certain type of visual pattern.
On the other hand, text query plays an important role in SGTL learning. We conduct experiments to replace text query which add conditional global textual features information with word query. The results show that simply using text query instead of word query has attained improvement by 0.28 \% on CUHK-PEDES and 0.56\% on ICFG-PEDES without CG and 0.19 \% on CUHK-PEDES and 1.19\% on ICFG-PEDES based on CG. Therefore, text query is important and essential for SGTL, especially for difficult datasets like ICFG-PEDES.

\subsection{Ablation Study of VGKT}

\paragraph{Comparison with Other Designs for VGKT}
\begin{table}[t]
\renewcommand\arraystretch{1.06}
	\centering
		\caption{\small{Comparison with other designs for VGKT on CUHK-PEDES and ICFG-PEDES datasets. FT, ST and CPT denote feature transfer, similarity transfer and class probability transfer, respectively.}}
    \begin{tabular}{cccc|cc|cc}
    \hline
    \multirow{2}{*}{Index}&	\multirow{2}{*}{FT}& 		\multirow{2}{*}{ST}& \multirow{2}{*}{CPT} & \multicolumn{2}{c|}{CUHK-PEDES} & \multicolumn{2}{c}{ICFG-PEDES} \\  
&		&	  &  & Rank-1 & Rank-5 &Rank-1  & Rank-5 \\ \hline  \hline 
1&		\xmarkg &    \xmarkg  & \xmarkg &  66.48& 83.64  & 58.85 &74.41  \\
2&		\cmark &    \xmarkg  & \xmarkg &    66.19  & 83.21 & 58.55  & 73.89   \\
3&		\xmarkg&	 \cmark  & \xmarkg  &   66.95 & 83.98  &  59.89   &  75.52   \\
4&		\xmarkg&	 \xmarkg  & \cmark   &  66.79  &  83.99 & 59.01   &  75.13   \\
5&		\xmarkg&	 \cmark  & \cmark    &   67.52 & 84.37  &   60.34   &  76.01\\
	\hline
	\end{tabular}

	\label{tab:VGKT-variations}
\end{table}

\begin{table*}[t]
\begin{center}
\renewcommand\arraystretch{1.06}
\footnotesize
\caption{Comparisons with state-of-the-art methods on the CUHK-PEDES. Best results are labeled in \textbf{bold}.}
\setlength{\tabcolsep}{2.9mm}
\begin{tabular}{r|c|c|c|c|c|c|c|c|c|c}
\hline
\multirow{2}{*}{\textbf{Method}} & \multirow{2}{*}{\textbf{Reference}} & \multirow{2}{*}{\textbf{Arch.}} & \multicolumn{4}{c|}{\textbf{Text to Image}}             & \multicolumn{4}{c}{\textbf{Image to Text}}             \\ \cline{4-11} 
                                 &        &                          & \textbf{Rank-1} & \textbf{Rank-5} & \textbf{Rank-10} & \textbf{mAP} & \textbf{Rank-1} & \textbf{Rank-5} & \textbf{Rank-10} & \textbf{mAP} \\ \hline \hline

GNA-RNN \cite{li2017person}     &CVPR’17                     & VGG16                          & 19.05           & -               & 53.64            & -            & -               & -               & -                & -            \\ 

Dual Path \cite{zheng2020dual}    &TOMM’20                     & ResNet50                           & 44.40           & 66.26           & 75.07            & -            & -               & -               & -                & -            \\ 

CMPM/C \cite{zhang2018cmpc} &ECCV’2018             & MobileNet                              & 49.37           & 71.69           & 79.27            & -        & 60.96           & 84.42           & 90.83            & -        \\ 
MIA \cite{niu2020mia}       &TIP’20                       & ResNet50                                & 53.10           & 75.00           & 82.90            & -            & -               & -               & -                & -            \\ 
PMA \cite{jing2020pose}    &AAAI’20                            & ResNet50                              & 54.12           & 75.45           & 82.97            & -            & -               & -               & -                & -            \\ 

TIMAM \cite{sarafianos2019adversarial}   &ICCV’19                         & ResNet101                               & 54.51           & 77.56           & 84.78            & -        & 67.40           & 88.65           & 93.91            & -        \\ 

CMKA \cite{chen2021cmka}          &TIP’21                   & ResNet50                                & 54.69           & 73.65           & 81.86            & -            & -               & -               & -                & -            \\ 
ViTAA \cite{wang2020vitaa}        &ECCV’20                   & ResNet50                               & 54.92           & 75.18           & 82.90            & 51.60        & 65.71           & 88.68           & 93.75            & 45.75        \\ 
CMAAM \cite{aggarwal2020cmaam}     &WACV’20                        & ResNet50                               & 56.68           & 77.18           & 84.86            & -            & -               & -               & -                & -            \\ 
HGAN \cite{zheng2020hierarchical}  &MM’20                         & ResNet50                                & 59.00           & 79.49           & 86.62            & -        & 71.16      & 90.05           & 95.06            & -        \\ 
NAFS (G) \cite{gao2021contextual}    &arXiv’21                           & ResNet50                            & 59.36           & 79.13          & 86.00            & 54.07            & 71.89               & 90.99               & 95.28                & 50.16            \\ 
MGEL \cite{wang2021mgel}       &IJCAI’21                       & ResNet50                          & 60.27           & 80.01          & 86.74            & -            & 71.87               & 91.38               & 95.42                & -            \\ 
AXM-Net \cite{farooq2021axm}     &AAAI’22                     & ResNet50                              & 61.90           & 79.41           & 85.75            & 57.38       & -               & -               & -                & -            \\ 
TIPCB \cite{chen2021tipcb}    &Neuro’22                      & ResNet50                            &63.63           & 82.82           & 89.01            & 56.78       & 73.55               & 92.26               & 96.03                & 51.78            \\ 
SSAN \cite{ssan}          &arXiv’21                 & ResNet50                                    & 61.37           & 80.15           & 86.73            & -       & -         & -          & -          & -      \\ 
TextReID~\cite{textreid}    &BMVC’21                          & ResNet50                                     & 61.65           & 80.98           & 86.78            & 58.29        & 75.96          & 93.40           & 96.55            & 55.05        \\ 
LapsCore~\cite{lapscore}     &ICCV’21               &      ResNet50   &                                   63.40&    -        & 87.80          &   -  &     -      &      -   &        -    &  -
\\ 
ACSA~\cite{ACSA}          &TMM’22                   &    Swin-Tiny                               &   63.56& 81.40& 87.70   &   -  &       -    &     -    &       -     & -
\\ 
ISANet~\cite{ISANet}          &arXiv’22                   &    ResNet50                               &   63.92& 82.15&         87.69    &   -  &       -    &     -    &       -     & -
\\ 
SRCF~\cite{SRCF}          &ECCV’22                   &    ResNet50                               &   64.04& 82.99& 88.81    &   -  &       -    &     -    &       -     & -
\\ 
LBUL~\cite{LBUL}          &MM’22                   &    ResNet50                              &   64.04 &82.66 &87.22    &   -  &       -    &     -    &       -     & -
\\ 

LGUR~\cite{lgur}          &MM’22                   &    ResNet50                               &   64.21         &            81.94&         87.93    &   -  &       -    &     -    &       -     & -
\\ 

CAIBC~\cite{CAIBC}          &MM’22                   &    ResNet50                              &   64.43 &82.87& 88.37    &   -  &       -    &     -    &       -     & -
\\ 
C$_2$A$_2$~\cite{C2A2}          &MM’22                   &    ResNet50                               &   64.82& 83.54 &89.77    &   -  &       -    &     -    &       -     & -
\\ 
IVT~\cite{IVT}          &ECCVW’22                   &    ViT-Base                               &   65.59         &            83.11&         89.21    &   -  &       -    &     -    &       -     & -
\\ 
CFine~\cite{CFine}          &arXiv’22                   &    ViT-Base                               &   69.57 & 85.93  &91.15   &   -  &       -    &     -    &       -     & -
\\

\hline

\textbf{\ourmodel}~(Ours)       &                      &    ResNet50                               &         67.52   &      84.37      &   90.26        &  63.67   &         82.49  &     95.64   &     97.78       & 61.42 \\
\textbf{\ourmodel}~(Ours)       &                      &    ViT-Base                               &         \textbf{71.38}& \textbf{86.75}  &\textbf{91.86}  & \textbf{67.91}  & \textbf{84.92}  &  \textbf{96.35} & \textbf{98.24} &\textbf{63.83}
\\ \hline
\end{tabular}
\label{tab:cuhk}
\end{center}
\end{table*}

\begin{figure*}[t]
	\centering
	\includegraphics[width = 0.9\textwidth]{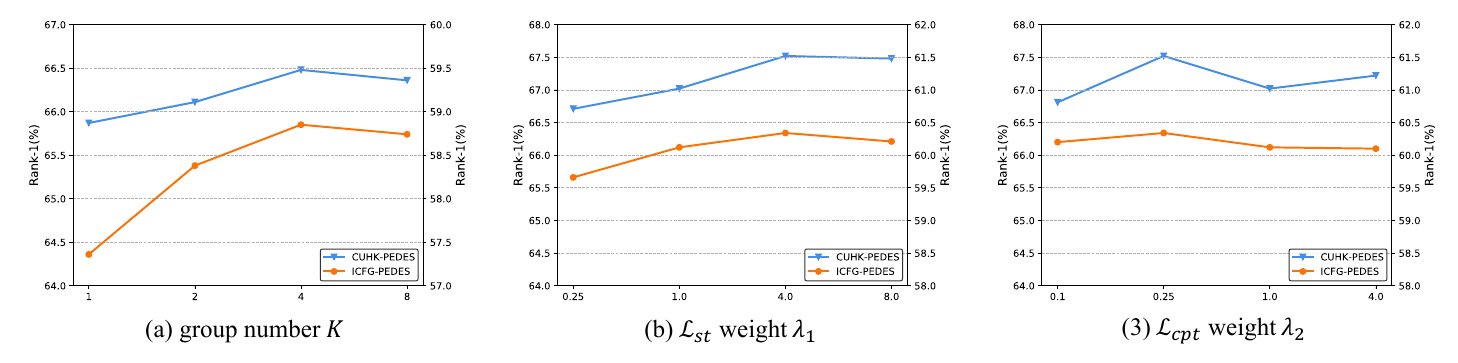}
	\caption{\small{Effects of three hyper-parameters, group number $K$, $\mathcal{L}_{st}$ weight $\lambda_1$, and $\mathcal{L}_{cpt}$ weight $\lambda_2$ on CUHK-PEDES and ICFG-PEDES datasets in terms of Rank-1 accuracy. 
}}
	\label{fig:hyper-parameter}
\end{figure*}

We further evaluate our model with our proposed VGKT. As shown in \tablename~\ref{tab:VGKT-variations}, when we take feature transfer to constrain numerical value between feature $VT_l^k$ and $T_l^k$, the performance drops by -0.29\% on CUHK-PEDES and -0.3\% on ICFG-PEDES in terms of Rank-1 accuracy, respectively (index 2). This confirms that simply narrowing the distance between two features like 
conventional knowledge distillation is useless and might lead to discriminative information loss. 
Compared with this straightforward approach, we use the dark knowledge of the relationship between samples to have better generalization ability.
By introducing similarity transfer, Rank-1 accuracy is significantly improved by 0.47\% on CUHK-PEDES and 1.04\% on ICFG-PEDES. This suggests that pairwise similarity between visual features and textual features can act as a strong cue for knowledge transfer because the similarity score between cross-modal features is critical for the final performance (index 3).
Then, we evaluate class probability transfer. It can bring in 0.31\% and 0.16 improvement in terms of Rank-1 accuracy on CUHK-PEDES and ICFG-PEDES, respectively (index 4). 
Finally, we apply for both similarity transfer and class probability transfer for the SGTL generation, and the performance has been further improved (index 5).
Therefore, similarity transfer and class probability transfer are complementary to each other, which is also consistent with the loss function including similarity-based contrastive loss and class probability-based instance loss.

\paragraph{Evaluation of the Hyper-parameter $\lambda_1$ and $\lambda_2$}
We evaluate the inference of the hyper-parameter $\lambda_1$ and $\lambda_2$ in the final loss function Eq.\ref{eq:loss_function}. As shown in \figurename~\ref{fig:hyper-parameter} (b-c), our hyper-parameter are relatively stable and there is no big fluctuation. We set $\lambda_1$ to 4.0 and $\lambda_2$ to 0.25.

\subsection{Comparison with State-of-the-Art Methods}

\begin{figure*}[t]
	\centering
	\includegraphics[width = 0.9\textwidth]{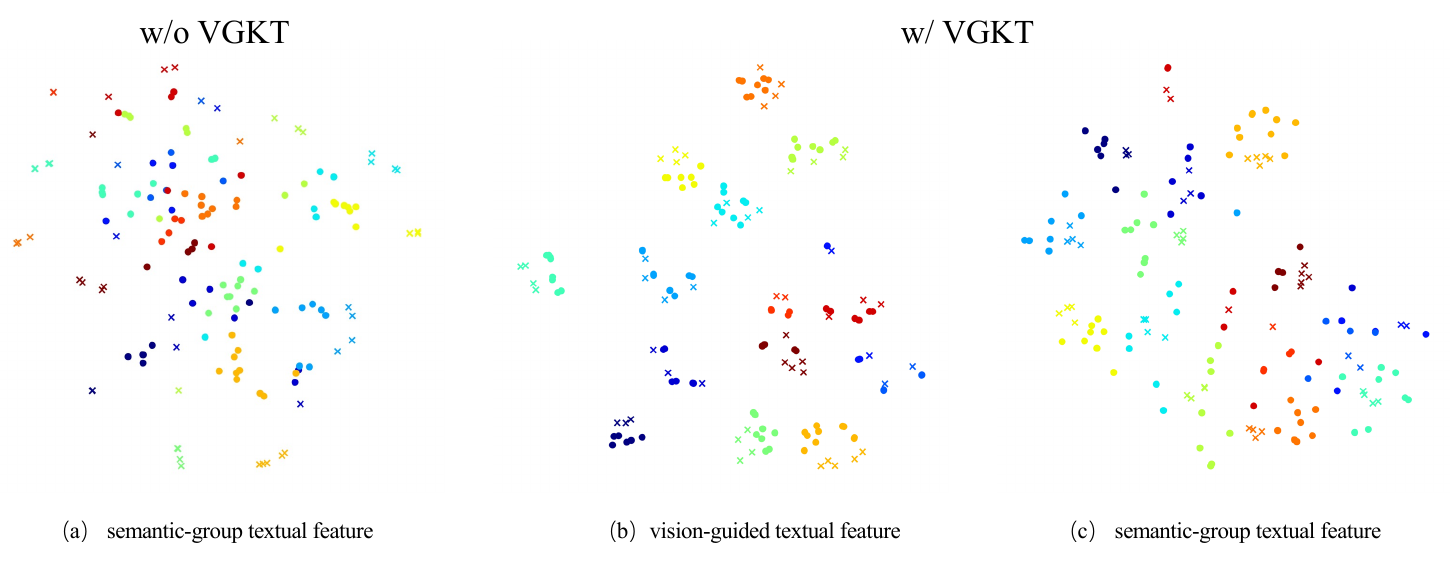}
	\caption{\small{t-SNE visualization on different kinds of the local visual features and local textual feature distribution. The circle denotes the local textual feature. The cross denotes the local visual feature. Different colors represent different classes. 
}}\label{fig:tsne}
\end{figure*}
\paragraph{Performance Comparison on CUHK-PEDES}
We compare \ourmodel~with state-of-the-art methods on CUHK-PEDES with text-to-image and image-to-text settings in \tablename~\ref{tab:cuhk} to show our superiority.
Our \ourmodel~outperforms the previous SOTA methods by a large margin with 2.70\%/1.81\% Rank-1 and 6.53\%/8.96\% Rank-1 for text-image and image-text using ResNet50/ViT, respectively.
Specifically, \ourmodel~(ResNet50) outperform TextReID~\cite{textreid} by 5.87\% Rank-1 and 6.53\% Rank-1 for text-image and image-text, respectively. It is worth noting that TextReID and \ourmodel~both utilize CLIP to extract visual and textual features. While TextReID does not make full use of CLIP and add redundant GRU to extract textual feature which leads to a performance gap. Furthermore, we apply the local branch to mine fine-grained information which is critical for retrieval tasks and neglected by TextReID. 

\begin{table}[t]
\renewcommand\arraystretch{1.06}
\centering
\footnotesize
\caption{\small{Comparisons with state-of-the-art methods on the ICFG-PEDES. Best results are labeled in \textbf{bold}.}}
\setlength{\tabcolsep}{1.5mm}
\begin{tabular}{r|c|c|ccc}
\hline
Method  &  Reference & Arch.  & Rank-1 &Rank-5 & Rank-10 \\ \hline \hline
Dual Path \cite{zheng2020dual}     &  TOMM’20 &ResNet50   & 38.99 & 59.44 &68.41      \\ 
CMPM/C \cite{zhang2018cmpc} & ECCV’18    &MobileNet       & 43.51 & 65.44 & 74.26               \\ 
MIA \cite{niu2020mia}   &    TIP’20   &ResNet50        & 46.49 & 67.14 & 75.18                   \\ 
SCAN~\cite{lee2018stacked}   &  ECCV’18  &ResNet50          & 50.05 & 69.65 & 77.21       \\ 
ViTAA \cite{wang2020vitaa}& ECCV’20  &ResNet50    &  50.98 & 68.79 & 75.78     \\ 
SSAN~\cite{ssan}  &  arXiv’21   &ResNet50     & 54.23 & 72.63 & 79.53          \\ 

IVT~\cite{IVT}     &ECCVW’22   &ViT-Base             &   56.04 & 73.60&  80.22 \\ 
SRCF~\cite{SRCF}          &ECCV’22 &ResNet50  &57.18 &75.01 &81.49
\\ 
LGUR~\cite{lgur}   &      MM’22     &ResNet50         &   57.42   &      74.97 &         81.45 \\

ISANet~\cite{ISANet}           &arXiv’22 &ResNet50 &57.73 &75.42&81.72\\
CFine~\cite{CFine}          &arXiv’22 &ViT-Base  &60.83 &76.55&82.42\\
\hline
\textbf{\ourmodel}~(Ours)      &                  &  ResNet50 & 60.34       &     76.01     & 82.01 \\
\textbf{\ourmodel}~(Ours)      &                  &  ViT-Base&  \textbf{63.05}        &     \textbf{78.43}       & \textbf{84.36} \\
\hline
\end{tabular}
\label{tab:icfg}
\end{table}

\begin{table}[t]
\renewcommand\arraystretch{1.06}
\begin{center}
\caption{\small{Comparison with state-of-the-art methods on Domain Generalization task. Best results are labeled in \textbf{bold}.\label{tab:dg}}}
\setlength{\tabcolsep}{3.96mm}{\begin{tabular}{c|r|ccc}
\hline
&Method  & Rank-1 &Rank-5 & Rank-10 \\ \hline \hline
\multirow{6}{*}{\rotatebox{90}{C $\rightarrow$ I}} & Dual Path \cite{zheng2020dual}           & 15.41& 29.80 &38.19      \\ 
&MIA \cite{niu2020mia}                 & 19.35  &36.78 & 46.42                  \\ 
&SCAN~\cite{lee2018stacked}               &21.27 &39.26 &48.83       \\ 

&SSAN~\cite{ssan}            & 24.72& 43.43 &53.01         \\ 

&LGUR~\cite{lgur}                     &   34.25& 52.58& 60.85 \\ \hline
&\textbf{\ourmodel}~(Ours)                       &      \textbf{35.85}      &     \textbf{55.04}       &\textbf{63.61}
\\ \hline
\multirow{6}{*}{\rotatebox{90}{I $\rightarrow$ C}} & Dual Path \cite{zheng2020dual}           & 7.63 &17.14& 23.52      \\ 
&MIA \cite{niu2020mia}                 & 10.93 &23.77 &32.39                  \\ 
&SCAN~\cite{lee2018stacked}               & 13.63 &28.61 &37.05       \\ 

&SSAN~\cite{ssan}            & 16.68& 33.84 &43.00         \\ 

&LGUR~\cite{lgur}                     &   25.44 &44.48& 54.39 \\ \hline
&\textbf{\ourmodel}~(Ours)                        &   \textbf{27.17}  &   \textbf{47.77}  & \textbf{57.27}
\\ \hline
\end{tabular}}
\end{center}

\end{table}

\paragraph{Performance Comparison on ICFG-PEDES}
We evaluate the performance of \ourmodel~with state-of-the-art methods on the other dataset named ICFG-PEDES. As shown in~\tablename~\ref{tab:icfg}, 
due to ICFG-PEDES being a new database, we directly compare the existing methods on text to image task. 
our method still achieves the best performance with 63.05\% Rank-1 accuracy and outperforms the previous SOTA method CFine~\cite{CFine} by 2.22\% in terms of Rank-1 accuracy.

\paragraph{Performance Comparison on the Domain Generalization Task}
As shown in~\tablename~\ref{tab:dg}, to further validate the superiority of our method, we conduct experiments on domain generalization task following~\cite{lgur}. We simply test the performance of model that is pretrained on the source domain on the target domain. Our \ourmodel~achieves the best performance among all the other methods. Specifically, we outperform LGUR~\cite{lgur} by 1.6\%, 1.73\% in terms of Rank-1 accuracy on C $\rightarrow$ I and I $\rightarrow$ C, respectively.
This experiment reveals that our local visual and textual alignment have good generalization capability.

\begin{table}[t!]
\renewcommand\arraystretch{1.06}
\centering
\caption{\small{Computational cost comparison among state-of-the-art methods on the CUHK-PEDES database. ``CAM'' represents the cross-modal attention mechanism.}}
\setlength{\tabcolsep}{3.36mm}{\begin{tabular}{c|r|cccc}
    \hline
       CAM & \multicolumn{1}{c|}{Methods}   & FLOPs & Inference & Rank-1 \\
    \hline
    \hline
     \cmark & MIA \cite{niu2020mia}  & 6.95G &13.1ms  &53.10\\
       \cmark & SCAN \cite{lee2018stacked}  &7.02G &13.7ms &55.86\\
       \cmark & NAFS \cite{gao2021contextual}  & 8.87G &15.2ms &59.94\\
       \xmarkg & Dual Path \cite{zheng2020dual}  & 2.92G &2.3ms & 44.40\\
       \xmarkg & CMPM/C \cite{zhang2018cmpc}  &3.11G &3.1ms &49.37\\
       \xmarkg & SSAN \cite{ssan}  & 6.94G &6.0ms &61.37\\
        \xmarkg & \ourmodel~(Ours) & 6.79G &5.3ms & 67.10 \\
    \hline
\end{tabular}}
\label{tab:costtime}
\end{table}

\subsection{Computational Complexity}

As mentioned in the introduction, we aim to design an efficient and effective pipeline for text-based person search and get rid of the pair-comparison paradigm. Thus, we have undertaken a series of experiments to assess the inference speed of each text-to-image query and the model complexity of various methods, employing a consistent and equitable experimental setup. Specifically, a batch size of 64 is employed, and all experiments are executed on a single Nvidia V100 GPU using the PyTorch toolkit. 
As shown in \tablename~\ref{tab:costtime}, the proposed \ourmodel~is faster than SSAN~\cite{ssan} by 0.13x on the CUHK-PEDES database. More notably, when compared to approaches reliant on the pair-comparison paradigm, such as MIA~\cite{niu2020mia}, SCAN~\cite{lee2018stacked}, and NAFS~\cite{gao2021contextual}, the proposed \ourmodel~ demonstrates substantially faster inference times and lower FLOPs. This achievement can be attributed to our utilization of knowledge distillation, a key factor in shedding the pair-comparison paradigm and attaining impressive results.
Collectively, the outcomes of these comparative analyses provide evidence that the proposed \ourmodel~stands as an effective and efficient pipeline for text-based person search.

\begin{figure*}[t]
	\centering
	\includegraphics[width = 0.976\textwidth]{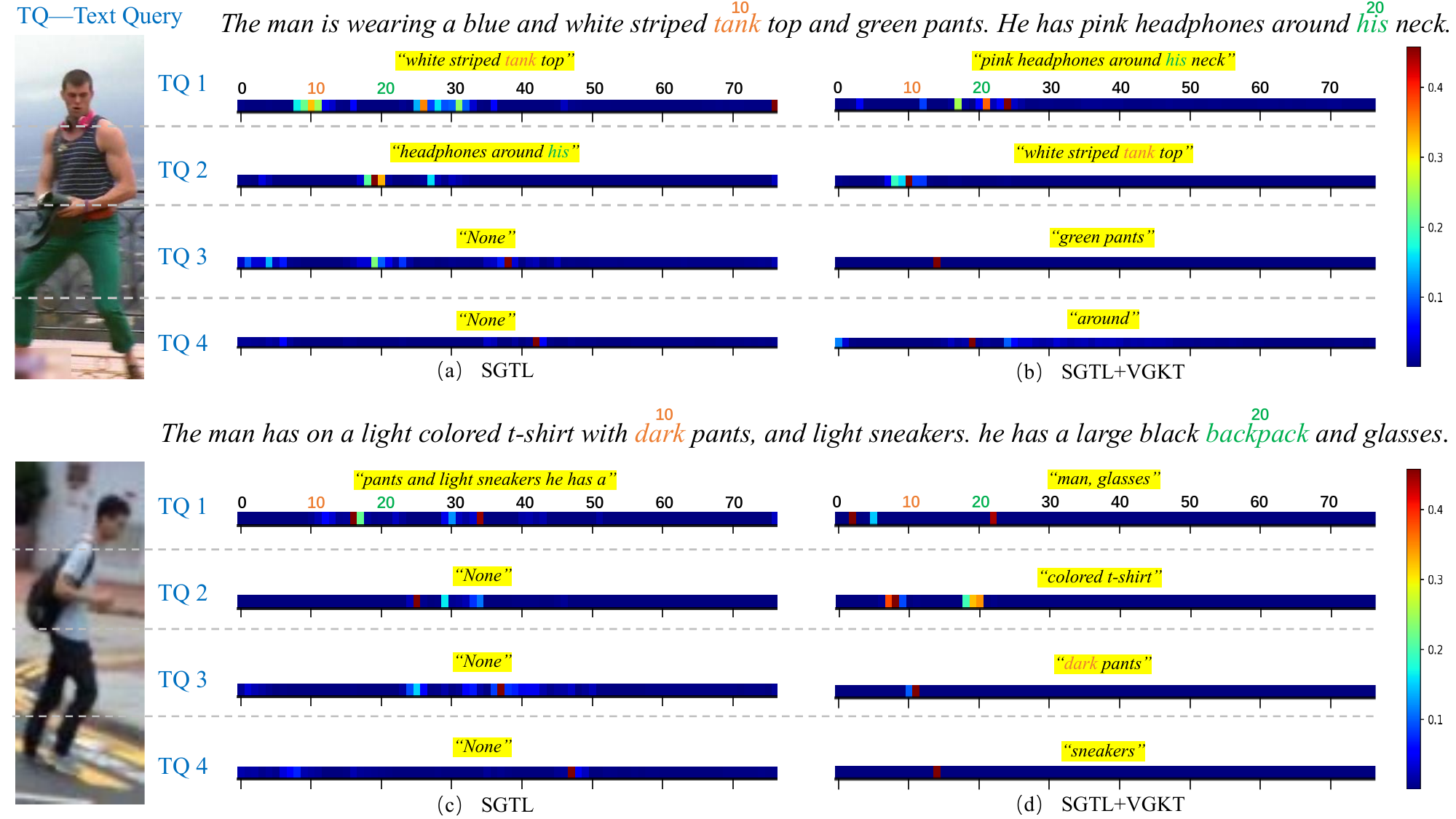}
	\caption{\small{Visualization of attention maps for four text queries within the SGTL cross-attention block. Each attention map represents the relationship between the $i$-th text query and the original text description, aiming to estimate its ability to grasp specific fixed-strip body parts. All sentences are standardized to a length of 77 words, with zero padding applied to sentences shorter than 77 words. 
    It is worth noting that the language information corresponding to zero padding is denoted as ``None''. For ease of comprehension, we have highlighted the 10-th word in orange and the 20-th word in green.
Additionally, words with significant attention weight are denoted with a yellow background color.
 Different colors represent different attention scores. Please refer to the colormap on the right for the correspondence between colors and attention values.
}}\label{fig:attention_visualize}
\end{figure*}

\subsection{Quantitative Result}

\paragraph{Visualization of Feature Distribution}

We utilize t-SNE to visualize the features distribution with and without VGKT in \figurename~\ref{fig:tsne} to verify our superiority of VGKT.
Note that we use paired local textual and visual features (here we use 1-th feature for simplicity) and take features from 15 classes on CUHK-PEDES test set. Local textual features are semantic-group textual features and vision-guided semantic-group textual features.
As shown in \figurename~\ref{fig:tsne}(a), semantic-group textual features from SGTL and corresponding local visual feature (such as the brown circles and brown cross) are far away from each other. 
In addition, inter-class distance is not large enough. The reason behind this is that SGTL is conducted alone from the language itself and has no interaction with the image. In addition, there is misalignment between semantic-group textual features and local visual features.
 
In \figurename~\ref{fig:tsne}(b), we visualize vision-guided textual features distribution which is perfectly aligned with local visual features and has high intra-class compactness and inter-class separability characteristics.
The phenomenon demonstrates that vision-guided textual features are discriminative and can serve as teacher distribution. In addition, vision-guided textual features are well aligned with local visual features.

In \figurename~\ref{fig:tsne}(c), 
since we transfer knowledge from vision-guided textual features to semantic-group textual features, semantic-group textual features are well aligned with local visual features. 
Consequently, the local textual features and local visual features of the same identity (such as the brown circles and brown cross) are pulled closer. Besides, the inter-class distance are larger than that in \figurename~\ref{fig:tsne}(a). The comparisons indicate that VGKT can effectively transfer these valuable knowledge from vision-guided textual features to semantic-group textual features.

\paragraph{Visualization of Attention Map}

To further verify the merits of our proposed method, we visualize the attention maps within the cross-attention block of SGTL, both with and without VGKT in  \figurename~\ref{fig:attention_visualize}. The left four rows are the fixed-stripe body parts and the right four rows are the attention maps representing the responses of the four text queries (TQ 1 to TQ4) with the whole language expression. 
As shown in \figurename~\ref{fig:attention_visualize}~(a) and (c), the four text queries attended areas of language are disorganized and have no correspondence with the local visual feature on the left.
The reason behind this is that SGTL is conducted alone from the language itself and has no interaction with the image.
In contrast, in \figurename~\ref{fig:attention_visualize}~(b) and (d), 
since we transfer knowledge from vision-guided textual features to semantic-group textual features, semantic-group textual features grasp the distribution of local visual features. 
Consequently, the $i$-th text query aligns with its corresponding local visual feature of the $i$-th stripe, such as the 1-th text query has high attention scores in the ``pink headphones around his neck'' corresponding to the 1-th local visual feature.  This phenomenon demonstrates that the proposed VGKT module can effectively transfer this crucial knowledge from vision-guided textual features to semantic-group textual features. As such, semantic-group textual features are well-aligned with local visual features.

\section{Conclusion}
We propose a Vision-Guided Semantic-Group network, termed VGSG, to mitigate the misalignment of fine-grained cross-modal features for Text-based Person Search without external tools or annotations. VGSG consists of a Semantic-group Textual Learning (SGTL) module and a Vision-guided Knowledge Transfer (VGKT) module. In SGTL, we group feature channels of textual features based on the semantic cues of language expression, so that features with similar semantic distribution are grouped together without extra tools. In VGKT, we adopt a vision-guided attention to excavate textual features relevant to visual concepts and term the output vision-guided textual features. To encourage getting rid of pairwise interaction and aligning with visual cues, we further transfer the relation knowledge from vision-guided textual features as teacher to the generation of semantic-group textual features.
Comprehensive experiments on two datasets demonstrate the superiority and effectiveness of our VGSG.

\ifCLASSOPTIONcaptionsoff

\fi

{\small
\bibliographystyle{IEEEtran}
\bibliography{reference}

\begin{thebibliography}{10}
\providecommand{\url}[1]{#1}
\csname url@samestyle\endcsname
\providecommand{\newblock}{\relax}
\providecommand{\bibinfo}[2]{#2}
\providecommand{\BIBentrySTDinterwordspacing}{\spaceskip=0pt\relax}
\providecommand{\BIBentryALTinterwordstretchfactor}{4}
\providecommand{\BIBentryALTinterwordspacing}{\spaceskip=\fontdimen2\font plus
\BIBentryALTinterwordstretchfactor\fontdimen3\font minus \fontdimen4\font\relax}
\providecommand{\BIBforeignlanguage}[2]{{%
\expandafter\ifx\csname l@#1\endcsname\relax
\typeout{** WARNING: IEEEtran.bst: No hyphenation pattern has been}%
\typeout{** loaded for the language `#1'. Using the pattern for}%
\typeout{** the default language instead.}%
\else
\language=\csname l@#1\endcsname
\fi
#2}}
\providecommand{\BIBdecl}{\relax}
\BIBdecl

\bibitem{li2017person}
S.~Li, T.~Xiao, H.~Li, B.~Zhou, D.~Yue, and X.~Wang, ``Person search with natural language description,'' in \emph{Proc. IEEE Conf. Comput. Vis. Pattern Recognit.}, 2017, pp. 1970--1979.

\bibitem{lgur}
Z.~Shao, X.~Zhang, M.~Fang, Z.~Lin, J.~Wang, and C.~Ding, ``Learning granularity-unified representations for text-to-image person re-identification,'' in \emph{ACM Int. Conf. Multimedia}, 2022, pp. 5566--5574.

\bibitem{lapscore}
Y.~Wu, Z.~Yan, X.~Han, G.~Li, C.~Zou, and S.~Cui, ``Lapscore: Language-guided person search via color reasoning,'' in \emph{Proc. IEEE Conf. Comput. Vis. Pattern Recognit.}, 2021, pp. 1624--1633.

\bibitem{textreid}
X.~Han, S.~He, L.~Zhang, and T.~Xiang, ``Text-based person search with limited data,'' in \emph{Proc. Brit. Mach. Vis. Conf.}, 2021, p. 337.

\bibitem{ssan}
Z.~Ding, C.~Ding, Z.~Shao, and D.~Tao, ``Semantically self-aligned network for text-to-image part-aware person re-identification,'' \emph{arXiv preprint arXiv:2107.12666}, 2021.

\bibitem{wang2021mgel}
C.~Wang, Z.~Luo, Y.~Lin, and S.~Li, ``Text-based person search via multi-granularity embedding learning,'' in \emph{Proc. Int. Joint Conf. Artif. Intell.}, 2021, pp. 1068--1074.

\bibitem{zheng2020hierarchical}
K.~Zheng, W.~Liu, J.~Liu, Z.~Zha, and T.~Mei, ``Hierarchical gumbel attention network for text-based person search,'' in \emph{ACM Int. Conf. Multimedia}, 2020, pp. 3441--3449.

\bibitem{gao2021contextual}
C.~Gao, G.~Cai, X.~Jiang, F.~Zheng, J.~Zhang, Y.~Gong, P.~Peng, X.~Guo, and X.~Sun, ``Contextual non-local alignment over full-scale representation for text-based person search,'' \emph{arXiv preprint}, 2021.

\bibitem{chen2021cmka}
Y.~Chen, R.~Huang, H.~Chang, C.~Tan, T.~Xue, and B.~Ma, ``Cross-modal knowledge adaptation for language-based person search,'' \emph{IEEE Trans. Image Process.}, vol.~30, pp. 4057--4069, 2021.

\bibitem{zheng2020dual}
Z.~Zheng, L.~Zheng, M.~Garrett, Y.~Yang, M.~Xu, and Y.-D. Shen, ``Dual-path convolutional image-text embeddings with instance loss,'' \emph{ACM Trans. Multimed. Comput. Commun. Appl.}, vol.~16, no.~2, pp. 1--23, 2020.

\bibitem{jing2020pose}
Y.~Jing, C.~Si, J.~Wang, W.~Wang, L.~Wang, and T.~Tan, ``Pose-guided multi-granularity attention network for text-based person search,'' in \emph{Proc. AAAI Conf. Artif. Intell.}, 2020, pp. 11\,189--11\,196.

\bibitem{farooq2021axm}
A.~Farooq, M.~Awais, J.~Kittler, and S.~S. Khalid, ``Axm-net: Cross-modal context sharing attention network for person re-id,'' \emph{arXiv preprint}, 2021.

\bibitem{chen2021tipcb}
Y.~Chen, G.~Zhang, Y.~Lu, Z.~Wang, Y.~Zheng, and R.~Wang, ``Tipcb: A simple but effective part-based convolutional baseline for text-based person search,'' \emph{Neurocomputing}, pp. 171--181, 2021.

\bibitem{li2017identity}
S.~Li, T.~Xiao, H.~Li, W.~Yang, and X.~Wang, ``Identity-aware textual-visual matching with latent co-attention,'' in \emph{Proc. IEEE Int. Conf. Comput. Vis.}, 2017, pp. 1890--1899.

\bibitem{VLT}
H.~Ding, C.~Liu, S.~Wang, and X.~Jiang, ``Vision-language transformer and query generation for referring segmentation,'' in \emph{Proc. IEEE Int. Conf. Comput. Vis.}, 2021, pp. 16\,321--16\,330.

\bibitem{aggarwal2020cmaam}
S.~Aggarwal, R.~V. Babu, and A.~Chakraborty, ``Text-based person search via attribute-aided matching,'' in \emph{Proc. IEEE Winter Conf. Appl. Comput. Vis.}, 2020, pp. 2606--2614.

\bibitem{loper2002nltk}
E.~Loper and S.~Bird, ``Nltk: The natural language toolkit,'' in \emph{Assoc. Comput. Linguist.}, 2004.

\bibitem{wang2020vitaa}
Z.~Wang, Z.~Fang, J.~Wang, and Y.~Yang, ``Vitaa: Visual-textual attributes alignment in person search by natural language,'' in \emph{Proc. Eur. Conf. Comput. Vis.}, 2020, pp. 402--420.

\bibitem{sun2018beyond}
Y.~Sun, L.~Zheng, Y.~Yang, Q.~Tian, and S.~Wang, ``Beyond part models: Person retrieval with refined part pooling (and a strong convolutional baseline),'' in \emph{Proc. Eur. Conf. Comput. Vis.}, 2018, pp. 480--496.

\bibitem{MGN}
G.~Wang, Y.~Yuan, X.~Chen, J.~Li, and X.~Zhou, ``Learning discriminative features with multiple granularities for person re-identification,'' in \emph{ACM Int. Conf. Multimedia}, 2018, pp. 274--282.

\bibitem{clip}
A.~Radford, J.~W. Kim, C.~Hallacy, A.~Ramesh, G.~Goh, S.~Agarwal, G.~Sastry, A.~Askell, P.~Mishkin, J.~Clark \emph{et~al.}, ``Learning transferable visual models from natural language supervision,'' in \emph{Proc. Int. Conf. Learn. Represent.}\hskip 1em plus 0.5em minus 0.4em\relax PMLR, 2021, pp. 8748--8763.

\bibitem{zhang2016picking}
X.~Zhang, H.~Xiong, W.~Zhou, W.~Lin, and Q.~Tian, ``Picking deep filter responses for fine-grained image recognition,'' in \emph{Proc. IEEE Conf. Comput. Vis. Pattern Recognit.}, 2016, pp. 1134--1142.

\bibitem{hinton2015distilling}
G.~Hinton, O.~Vinyals, J.~Dean \emph{et~al.}, ``Distilling the knowledge in a neural network,'' \emph{arXiv preprint arXiv:1503.02531}, 2015.

\bibitem{chen2018improving}
T.~Chen, C.~Xu, and J.~Luo, ``Improving text-based person search by spatial matching and adaptive threshold,'' in \emph{Proc. IEEE Winter Conf. Appl. Comput. Vis.}, 2018, pp. 1879--1887.

\bibitem{zhang2018cmpc}
Y.~Zhang and H.~Lu, ``Deep cross-modal projection learning for image-text matching,'' in \emph{Proc. Eur. Conf. Comput. Vis.}, 2018, pp. 707--723.

\bibitem{he2023region}
S.~He, W.~Chen, K.~Wang, H.~Luo, F.~Wang, W.~Jiang, and H.~Ding, ``Region generation and assessment network for occluded person re-identification,'' \emph{IEEE Trans. Inf. Forensics Secur.}, 2023.

\bibitem{R11}
X.~Chen, W.~Liu, X.~Liu, Y.~Zhang, and T.~Mei, ``A cross-modality and progressive person search system,'' in \emph{ACM Int. Conf. Multimedia}, 2020, pp. 4550--4552.

\bibitem{milnce}
A.~Miech, J.-B. Alayrac, L.~Smaira, I.~Laptev, J.~Sivic, and A.~Zisserman, ``End-to-end learning of visual representations from uncurated instructional videos,'' in \emph{Proc. IEEE Conf. Comput. Vis. Pattern Recognit.}, 2020, pp. 9879--9889.

\bibitem{simvlm}
Z.~Wang, J.~Yu, A.~W. Yu, Z.~Dai, Y.~Tsvetkov, and Y.~Cao, ``Simvlm: Simple visual language model pretraining with weak supervision,'' in \emph{Proc. Int. Conf. Learn. Represent.}, 2022.

\bibitem{howto100m}
A.~Miech, D.~Zhukov, J.-B. Alayrac, M.~Tapaswi, I.~Laptev, and J.~Sivic, ``Howto100m: Learning a text-video embedding by watching hundred million narrated video clips,'' in \emph{Proc. IEEE Conf. Comput. Vis. Pattern Recognit.}, 2019, pp. 2630--2640.

\bibitem{hairclip}
T.~Wei, D.~Chen, W.~Zhou, J.~Liao, Z.~Tan, L.~Yuan, W.~Zhang, and N.~Yu, ``Hairclip: Design your hair by text and reference image,'' in \emph{Proc. IEEE Conf. Comput. Vis. Pattern Recognit.}, 2022, pp. 18\,072--18\,081.

\bibitem{clip2video}
H.~Fang, P.~Xiong, L.~Xu, and Y.~Chen, ``Clip2video: Mastering video-text retrieval via image clip,'' \emph{arXiv preprint arXiv:2106.11097}, 2021.

\bibitem{clip4caption}
M.~Tang, Z.~Wang, Z.~Liu, F.~Rao, D.~Li, and X.~Li, ``Clip4caption: Clip for video caption,'' in \emph{ACM Int. Conf. Multimedia}, 2021, pp. 4858--4862.

\bibitem{PADing}
S.~He, H.~Ding, and W.~Jiang, ``Primitive generation and semantic-related alignment for universal zero-shot segmentation,'' in \emph{Proc. IEEE Conf. Comput. Vis. Pattern Recognit.}, 2023, pp. 11\,238--11\,247.

\bibitem{3DZero}
S.~He, X.~Jiang, W.~Jiang, and H.~Ding, ``Prototype adaption and projection for few- and zero-shot 3d point cloud semantic segmentation,'' \emph{IEEE Trans. Image Process.}, vol.~32, pp. 3199--3211, 2023.

\bibitem{D2Zero}
S.~He, H.~Ding, and W.~Jiang, ``Semantic-promoted debiasing and background disambiguation for zero-shot instance segmentation,'' in \emph{Proc. IEEE Conf. Comput. Vis. Pattern Recognit.}, 2023, pp. 19\,498--19\,507.

\bibitem{cho2014gru}
K.~Cho, B.~van Merri{\"e}nboer, C.~Gulcehre, D.~Bahdanau, F.~Bougares, H.~Schwenk, and Y.~Bengio, ``Learning phrase representations using {RNN} encoder{--}decoder for statistical machine translation,'' in \emph{{Proc. of the Conf. on Empirical Methods in Natural Language Process.}}, 2014, pp. 1724--1734.

\bibitem{vaswani2017transformer}
A.~Vaswani, N.~Shazeer, N.~Parmar, J.~Uszkoreit, L.~Jones, A.~N. Gomez, L.~Kaiser, and I.~Polosukhin, ``Attention is all you need,'' in \emph{Proc. Adv. Neural Inform. Process. Syst.}, 2017, pp. 5998--6008.

\bibitem{ren2019fastspeech}
Y.~Ren, Y.~Ruan, X.~Tan, T.~Qin, S.~Zhao, Z.~Zhao, and T.-Y. Liu, ``Fastspeech: Fast, robust and controllable text to speech,'' in \emph{Proc. Adv. Neural Inform. Process. Syst.}, 2019, pp. 3165--3174.

\bibitem{devlin2018bert}
J.~Devlin, M.-W. Chang, K.~Lee, and K.~Toutanova, ``{BERT}: Pre-training of deep bidirectional transformers for language understanding,'' in \emph{NAACL-HLT}, 2019, pp. 4171--4186.

\bibitem{VLTTPAMI}
H.~Ding, C.~Liu, S.~Wang, and X.~Jiang, ``{VLT}: Vision-language transformer and query generation for referring segmentation,'' \emph{IEEE Trans. Pattern Anal. Mach. Intell.}, vol.~45, no.~6, pp. 7900--7916, 2023.

\bibitem{GRES}
C.~Liu, H.~Ding, and X.~Jiang, ``{GRES}: Generalized referring expression segmentation,'' in \emph{Proc. IEEE Conf. Comput. Vis. Pattern Recognit.}, 2023, pp. 23\,592--23\,601.

\bibitem{MeViS}
H.~Ding, C.~Liu, S.~He, X.~Jiang, and C.~C. Loy, ``{MeViS}: A large-scale benchmark for video segmentation with motion expressions,'' in \emph{Proc. IEEE Int. Conf. Comput. Vis.}, 2023, pp. 2694--2703.

\bibitem{detr}
N.~Carion, F.~Massa, G.~Synnaeve, N.~Usunier, A.~Kirillov, and S.~Zagoruyko, ``End-to-end object detection with transformers,'' in \emph{Proc. Eur. Conf. Comput. Vis.}, 2020, pp. 213--229.

\bibitem{strudel2021segmenter}
R.~Strudel, R.~Garcia, I.~Laptev, and C.~Schmid, ``Segmenter: Transformer for semantic segmentation,'' in \emph{Proc. IEEE Conf. Comput. Vis. Pattern Recognit.}, 2021, pp. 7262--7272.

\bibitem{ding2018context}
H.~Ding, X.~Jiang, B.~Shuai, A.~Q. Liu, and G.~Wang, ``Context contrasted feature and gated multi-scale aggregation for scene segmentation,'' in \emph{Proc. IEEE Conf. Comput. Vis. Pattern Recognit.}, 2018, pp. 2393--2402.

\bibitem{vit}
A.~Dosovitskiy, L.~Beyer, A.~Kolesnikov, D.~Weissenborn, X.~Zhai, T.~Unterthiner, M.~Dehghani, M.~Minderer, G.~Heigold, S.~Gelly \emph{et~al.}, ``An image is worth 16x16 words: Transformers for image recognition at scale,'' in \emph{Proc. Int. Conf. Learn. Represent.}, 2020.

\bibitem{transreid}
S.~He, H.~Luo, P.~Wang, F.~Wang, H.~Li, and W.~Jiang, ``Transreid: Transformer-based object re-identification,'' in \emph{Proc. IEEE Conf. Comput. Vis. Pattern Recognit.}, 2021, pp. 15\,013--15\,022.

\bibitem{MOSE}
H.~Ding, C.~Liu, S.~He, X.~Jiang, P.~H. Torr, and S.~Bai, ``{MOSE}: A new dataset for video object segmentation in complex scenes,'' in \emph{Proceedings of the IEEE/CVF International Conference on Computer Vision (ICCV)}, 2023, pp. 20\,224--20\,234.

\bibitem{parisotto2015actor}
E.~Parisotto, J.~L. Ba, and R.~Salakhutdinov, ``Actor-mimic: Deep multitask and transfer reinforcement learning,'' in \emph{Proc. Int. Conf. Learn. Represent.}, 2016.

\bibitem{romero2014fitnets}
A.~Romero, N.~Ballas, S.~E. Kahou, A.~Chassang, C.~Gatta, and Y.~Bengio, ``Fitnets: Hints for thin deep nets,'' in \emph{Proc. Int. Conf. Learn. Represent.}, 2015.

\bibitem{chen2018darkrank}
Y.~Chen, N.~Wang, and Z.~Zhang, ``Darkrank: Accelerating deep metric learning via cross sample similarities transfer,'' in \emph{Proc. AAAI Conf. Artif. Intell.}, 2018, pp. 2852--2859.

\bibitem{R12}
X.~Chen, X.~Liu, W.~Liu, X.-P. Zhang, Y.~Zhang, and T.~Mei, ``Explainable person re-identification with attribute-guided metric distillation,'' in \emph{Proc. IEEE Conf. Comput. Vis. Pattern Recognit.}, 2021, pp. 11\,813--11\,822.

\bibitem{R21}
S.~Zhang, M.~Chen, J.~Chen, Y.-F. Li, Y.~Wu, M.~Li, and C.~Zhu, ``Combining cross-modal knowledge transfer and semi-supervised learning for speech emotion recognition,'' \emph{Knowledge-Based Syst.}, vol. 229, p. 107340, 2021.

\bibitem{he2016resnet}
K.~He, X.~Zhang, S.~Ren, and J.~Sun, ``Deep residual learning for image recognition,'' in \emph{Proc. IEEE Conf. Comput. Vis. Pattern Recognit.}, 2016, pp. 770--778.

\bibitem{cao2007learning}
Z.~Cao, T.~Qin, T.-Y. Liu, M.-F. Tsai, and H.~Li, ``Learning to rank: from pairwise approach to listwise approach,'' in \emph{Proc. Int. Conf. Mach. Learn.}, 2007, pp. 129--136.

\bibitem{xia2008listwise}
F.~Xia, T.-Y. Liu, J.~Wang, W.~Zhang, and H.~Li, ``Listwise approach to learning to rank: theory and algorithm,'' in \emph{Proc. Int. Conf. Mach. Learn.}, 2008, pp. 1192--1199.

\bibitem{MSMT17}
L.~Wei, S.~Zhang, W.~Gao, and Q.~Tian, ``Person transfer gan to bridge domain gap for person re-identification,'' in \emph{Proc. IEEE Conf. Comput. Vis. Pattern Recognit.}, 2018, pp. 79--88.

\bibitem{random_erase3}
Z.~Zhong, L.~Zheng, G.~Kang, S.~Li, and Y.~Yang, ``Random erasing data augmentation,'' in \emph{AAAI}, 2020, pp. 13\,001--13\,008.

\bibitem{stoc_depth}
G.~Huang, Y.~Sun, Z.~Liu, D.~Sedra, and K.~Q. Weinberger, ``Deep networks with stochastic depth,'' in \emph{Proc. Eur. Conf. Comput. Vis.}, 2016, pp. 646--661.

\bibitem{niu2020mia}
K.~Niu, Y.~Huang, W.~Ouyang, and L.~Wang, ``Improving description-based person re-identification by multi-granularity image-text alignments,'' \emph{IEEE Trans. Image Process.}, pp. 5542--5556, 2020.

\bibitem{sarafianos2019adversarial}
N.~Sarafianos, X.~Xu, and I.~A. Kakadiaris, ``Adversarial representation learning for text-to-image matching,'' in \emph{Proc. IEEE Int. Conf. Comput. Vis.}, 2019, pp. 5813--5823.

\bibitem{ACSA}
Z.~Ji, J.~Hu, D.~Liu, L.~Y. Wu, and Y.~Zhao, ``Asymmetric cross-scale alignment for text-based person search,'' \emph{IEEE Trans. Multimedia}, 2022.

\bibitem{ISANet}
S.~Yan, H.~Tang, L.~Zhang, and J.~Tang, ``Image-specific information suppression and implicit local alignment for text-based person search,'' \emph{arXiv preprint arXiv:2208.14365}, 2022.

\bibitem{SRCF}
W.~Suo, M.~Sun, K.~Niu, Y.~Gao, P.~Wang, Y.~Zhang, and Q.~Wu, ``A simple and robust correlation filtering method for text-based person search,'' in \emph{Proc. Eur. Conf. Comput. Vis.}, 2022, pp. 726--742.

\bibitem{LBUL}
Z.~Wang, A.~Zhu, J.~Xue, X.~Wan, C.~Liu, T.~Wang, and Y.~Li, ``Look before you leap: Improving text-based person retrieval by learning a consistent cross-modal common manifold,'' in \emph{ACM Int. Conf. Multimedia}, 2022, pp. 1984--1992.

\bibitem{CAIBC}
Z.~Wang, A.~Zhu, J.~Xue, X.~Wan, C.~Liu, T.~Wang, and Y.~Li\textcolor{white}{.}\!, ``Caibc: Capturing all-round information beyond color for text-based person retrieval,'' in \emph{ACM Int. Conf. Multimedia}, 2022, pp. 5314--5322.

\bibitem{C2A2}
K.~Niu, L.~Huang, Y.~Huang, P.~Wang, L.~Wang, and Y.~Zhang, ``Cross-modal co-occurrence attributes alignments for person search by language,'' in \emph{ACM Int. Conf. Multimedia}, 2022, pp. 4426--4434.

\bibitem{IVT}
X.~Shu, W.~Wen, H.~Wu, K.~Chen, Y.~Song, R.~Qiao, B.~Ren, and X.~Wang, ``See finer, see more: Implicit modality alignment for text-based person retrieval,'' in \emph{Proc. Eur. Conf. Comput. Vis.}, 2022, pp. 624--641.

\bibitem{CFine}
S.~Yan, N.~Dong, L.~Zhang, and J.~Tang, ``Clip-driven fine-grained text-image person re-identification,'' \emph{arXiv preprint arXiv:2210.10276}, 2022.

\bibitem{lee2018stacked}
K.-H. Lee, X.~Chen, G.~Hua, H.~Hu, and X.~He, ``Stacked cross attention for image-text matching,'' in \emph{Proc. Eur. Conf. Comput. Vis.}, 2018, pp. 201--216.

\end{thebibliography}
}

\vfill

\end{document}